\documentclass[11pt]{article}

\usepackage[preprint]{acl}

\usepackage{times}
\usepackage{latexsym}

\usepackage[T1]{fontenc}

\usepackage[utf8]{inputenc}

\usepackage{microtype}

\usepackage{inconsolata}

\usepackage{graphicx}

\usepackage{latexsym}

\usepackage{booktabs}
\usepackage{multirow}
\usepackage{amsmath}
\usepackage{amssymb}
\usepackage{amsthm}
\usepackage{graphicx}
\usepackage{url}
\usepackage{xcolor}
\usepackage{array}
\usepackage{makecell}
\usepackage{paralist}
\usepackage{placeins}
\usepackage{listings}
\lstdefinestyle{prompt}{
  basicstyle=\ttfamily\footnotesize,
  breaklines=true,
  breakatwhitespace=true,
  columns=flexible,
  showstringspaces=false,
  keepspaces=true,
  upquote=true,
  frame=single,
  framerule=0.4pt,
  framesep=4pt,
  xleftmargin=4pt,
  xrightmargin=4pt,
  literate=%
    {→}{{$\rightarrow$}}{1}%
    {—}{{---}}{1}%
    {…}{{\ldots}}{1}%
    {≥}{{$\geq$}}{1},
}

\usepackage{tcolorbox}
\tcbuselibrary{skins,breakable,listings}
\newtcblisting{promptlisting}[1][]{%
  enhanced,
  colback=gray!6,
  colframe=black!60,
  boxrule=0.5pt,
  arc=2.5mm,
  left=2mm, right=2mm, top=1.3mm, bottom=1.3mm,
  title=#1,
  fonttitle=\bfseries\sffamily\small,
  coltitle=white,
  colbacktitle=black!70,
  attach boxed title to top left={xshift=4mm,yshift=-2.5mm},
  boxed title style={
    colback=black!70, colframe=black!70,
    boxrule=0pt, arc=1.5mm,
    left=4pt, right=4pt, top=1pt, bottom=1pt,
  },
  before skip=6pt, after skip=5pt,
  listing only,
  listing options={
    style=prompt,
    basicstyle=\ttfamily\fontsize{9}{10}\selectfont,
    frame=none,
    backgroundcolor=,
    framesep=0pt,
    xleftmargin=0pt,
    xrightmargin=0pt,
  },
}

\usepackage{multicol}

\lstdefinestyle{promptcompact}{
  basicstyle=\ttfamily\scriptsize,
  breaklines=true,
  breakatwhitespace=true,
  columns=flexible,
  showstringspaces=false,
  keepspaces=true,
  upquote=true,
  frame=none,
  framesep=0pt,
  xleftmargin=0pt,
  xrightmargin=0pt,
  aboveskip=1pt,
  belowskip=1pt,
  literate=%
    {→}{{$\rightarrow$}}{1}%
    {—}{{---}}{1}%
    {…}{{\ldots}}{1}%
    {≥}{{$\geq$}}{1},
}

\newtcolorbox{promptboxcompact}[1][]{%
  enhanced,
  colback=gray!6,
  colframe=black!60,
  boxrule=0.5pt,
  arc=2.5mm,
  left=3mm, right=3mm, top=2mm, bottom=2mm,
  title=#1,
  fonttitle=\bfseries\sffamily\small,
  coltitle=white,
  colbacktitle=black!70,
  attach boxed title to top left={xshift=4mm,yshift=-2.5mm},
  boxed title style={
    colback=black!70, colframe=black!70,
    boxrule=0pt, arc=1.5mm,
    left=4pt, right=4pt, top=1pt, bottom=1pt,
  },
  before skip=8pt, after skip=6pt,
}

\theoremstyle{plain}
\newtheorem{proposition}{Proposition}

\newcommand{\Levo}{L_{\text{evo}}}
\newcommand{\Ladapt}{L_{\text{adapt}}}
\newcommand{\Hist}{\mathcal{H}_t}

\newcommand{\zw}[1]{#1}  

\title{Adaptive Auto-Harness: Sustained Self-Improvement for Agentic System Deployment on Open-Ended Task Streams}

\usepackage[boxed]{amazontitle}
\amazontitlesetup{
  author-layout={2,1,1},
  title-break-after={:},
  logo={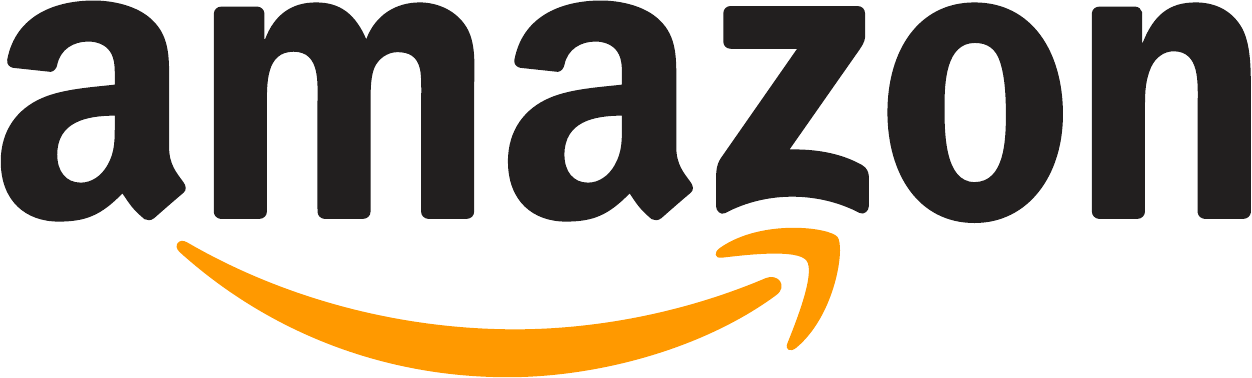},
  correspondence={\amazontitleemail{luhanqin@amazon.com}}
}

\author{Zewen Liu$^{1}$, Zhan Shi$^{2}$, Yisi Sang$^{2}$, Bing He$^{2}$, Minhua Lin$^{3}$, Tianxin Wei$^{4}$\\
\textbf{Dakuo Wang}$^{5}$, \textbf{Benoit Dumoulin}$^{2}$, \textbf{Wei Jin}$^{1}$, \textbf{Hanqing Lu}$^{2}$\\
$^{1}$Emory University $^{2}$Amazon $^{3}$The Pennsylvania State University $^{4}$UIUC $^{5}$Northeastern University \\
\texttt{\{zewen.liu,wei.jin\}@emory.edu}; 
{luhanqin@amazon.com}
}

\begin{document}
\maketitle
\suppressfloats[t]

\begin{abstract}
\sloppy\def\UrlBreaks{\do\/\do\-}
Auto-harness systems improve LLM agents by adapting prompts, skills, tools, and supporting infrastructure from execution feedback. Sustaining these gains on open-ended task streams requires handling growing histories, heterogeneous tasks, and shifting distributions. We introduce Adaptive Auto-Harness, a unified framework that separates persistent evolution workspaces from deployable harnesses and supports interchangeable evolution procedures. We compare configurations adapted from five prior approaches on recorded streams of prediction markets, security challenges, and event forecasts. With the same Sonnet solver, our full system reaches 80.9\% market all-task accuracy and 50.2\% security Pass@1, versus 22.2\% and 37.2\% without evolution; market accuracy also exceeds a market-following control's 72.0\%. Our analyses show that artifact accumulation does not guarantee sustained gains, and knowledge that transfers across tasks can still lose value as the task distribution changes. Continued evolution can therefore help preserve earlier gains through revision of accumulated knowledge. Code: \href{https://github.com/A-EVO-Lab/a-evolve}{\nolinkurl{https://github.com/A-EVO-Lab/a-evolve}}.
\end{abstract}

\section{Introduction}
\label{sec:intro}

\begin{figure}[t]
\centering
\includegraphics[width=\linewidth]{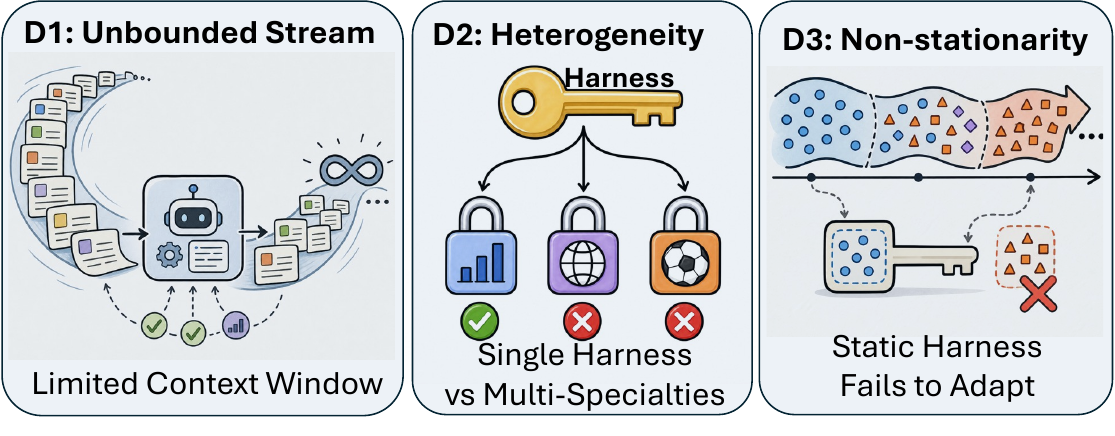}
\caption{\textbf{Task-stream deployment.}
Histories grow, tasks have different requirements, and distributions change
over time.}
\label{fig:teaser}
\end{figure}

Many real-world services handle ongoing streams of tasks, from
forecasting questions~\cite{halawi2024forecasting} to software maintenance tasks~\cite{jimenez2024swe} and security
investigations~\cite{deng2024pentestgpt}. An LLM agent deployed in such a service encounters an
\emph{open-ended task stream}: new tasks arrive without a predetermined
deployment endpoint, while execution records and feedback
accumulate from earlier tasks~\cite{zheng2025lifelongagentbench}. From accumulated task experience, the agent
can extract knowledge about the task domain, solution strategies, and tool
use~\cite{zhao2024expel,wang2023voyager}. Related tasks offer opportunities to apply retained knowledge and
avoid repeating earlier mistakes~\cite{shinn2023reflexion}. Experience reuse requires retaining
useful knowledge in a form that can guide the agent on subsequent tasks.

\emph{Harness evolution} supports experience reuse on subsequent tasks. An agent's
harness comprises the prompts, skills, tools, memory, and supporting
infrastructure surrounding its LLM. Auto-harness systems such as
A-Evolve~\cite{lin2026position}, GEPA~\cite{agrawal2025gepa}, and
Meta-Harness~\cite{lee2026meta} use task experience to update the harness
without updating model weights. Each harness update can encode knowledge
from task experience in skills, tools, or instructions for subsequent tasks.
Reported gains on offline benchmarks such as
SWE-bench~\cite{jimenez2024swe} provide evidence for experience reuse through
harness evolution: harness updates based on task experience can improve
performance on held-out tasks.

On ongoing task streams, harness evolution faces three deployment
challenges~\cite{karten2026continual,wang2023voyager}
(Figure~\ref{fig:teaser}). \textbf{(D1) Unbounded streams} require distilling
growing histories within a finite working context.
\textbf{(D2) Task heterogeneity} means knowledge useful for one task may
not suit another. \textbf{(D3) Distributional non-stationarity} can make
earlier knowledge less useful as task distributions change.
Each update must integrate new experience, retain useful knowledge, and
revise outdated knowledge, while the solver needs to apply knowledge
relevant to the incoming task.

\begin{figure}[t]
\centering
\includegraphics[width=\linewidth]{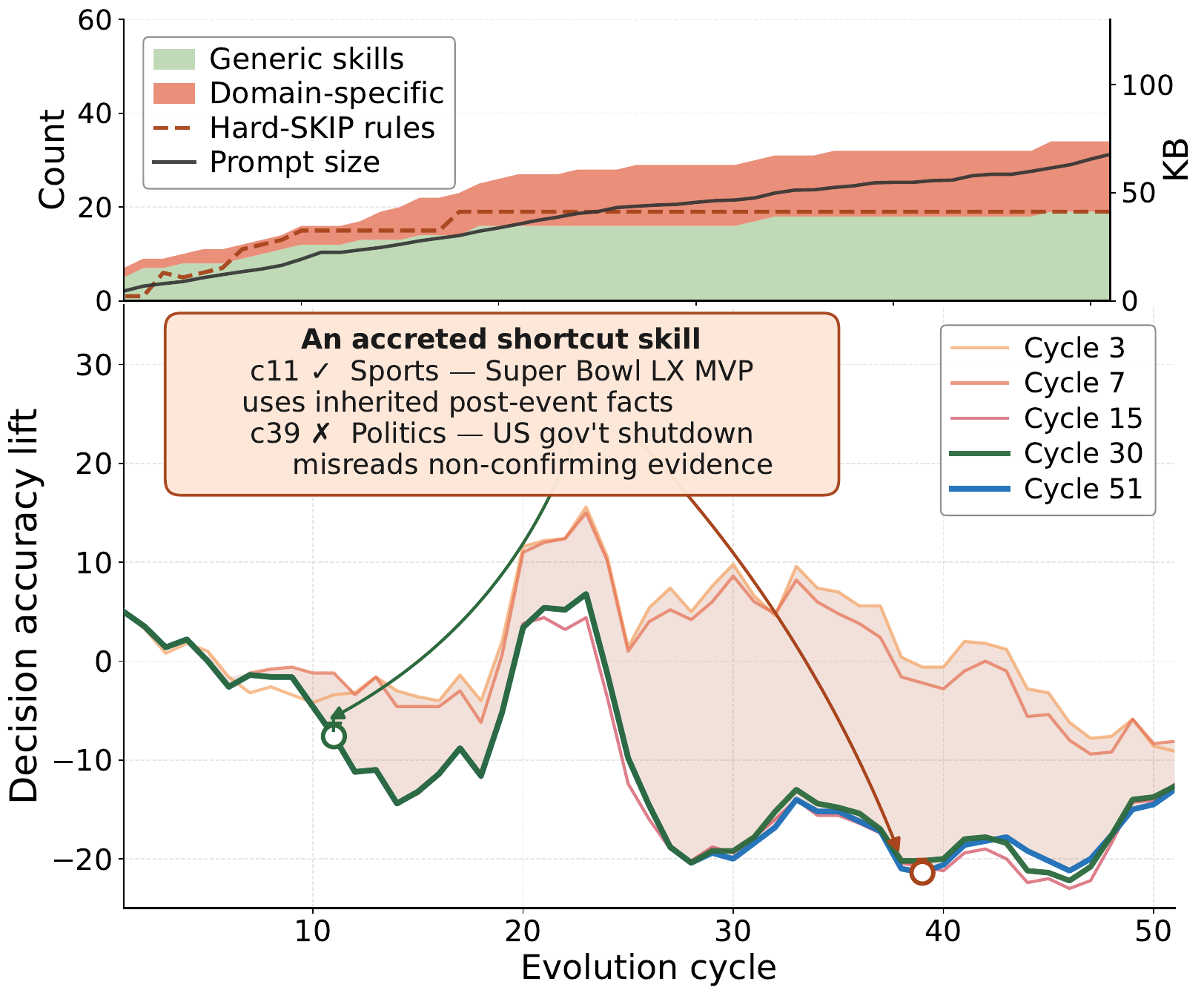}
\caption{\textbf{Artifact growth and subsequent performance.}
A recorded A-Evolve history and four frozen continuations.
\textbf{Top:} skill count, prompt size, and hard-\textsc{skip} rules.
\textbf{Bottom:} all-task accuracy lift over a static solver
(points; centered five-cycle means).}
\label{fig:c1}
\end{figure}

\begin{samepage}
Continued harness evolution can therefore expand the harness without
yielding consistent gains on subsequent tasks. Figure~\ref{fig:c1}
illustrates the gap in a recorded history of harness evolution: prompts
and skill collections grow across updates, while performance on subsequent
tasks does not improve consistently. We ask:
\textbf{Does harness evolution turn accumulated experience into reusable
knowledge on ongoing task streams?}
\par
\end{samepage}

We develop \textbf{Adaptive Auto-Harness}, a unified framework for maintaining
and applying knowledge on task streams. A common execution layer connects
the task stream, solver, and deployable harness to interchangeable evolution
procedures. We adapt evolution components from prior auto-harness approaches
to this interface and implement a stateful multi-agent configuration.
Our study covers prediction
markets~\cite{cheng2026polybench}, security
challenges~\cite{zhuo2025training}, and event
forecasting~\cite{zeng2025futurex}. We compare evolving and fixed harnesses
on finite stream prefixes, then examine artifact changes, solver behavior,
and optional deployment mechanisms. The unit of comparison is an evolution
configuration within the shared framework; implementation details and
information conditions are specified with the experiments.

Our contributions are:
\begin{compactenum}[(1)]
  \item \textbf{Task-stream evaluation for harness evolution.}
  We formulate experience reuse through harness evolution as a task-stream
  evaluation problem and construct evaluation streams in three domains
  from existing benchmarks. The setup follows successive harness updates
  and performance on subsequent tasks as execution histories accumulate.
  \item \textbf{A shared framework for evolution and deployment.}
  Adaptive Auto-Harness separates knowledge maintenance across updates
  from harness deployment on individual tasks. It supports different
  evolution procedures through a common interface, including our stateful
  multi-agent configuration. Persistent research records and versioned
  harnesses support implementation, comparison, and inspection of
  experience reuse.
  \item \textbf{Empirical insights into sustained improvement.}
  Evolution can improve performance with fixed solver weights, but larger
  artifact collections do not ensure sustained gains. Behavior diagnostics
  expose remaining capability bottlenecks. Frozen-harness replays further
  link larger distribution shifts to greater losses relative to continued
  evolution, highlighting knowledge revision as a way to preserve earlier
  gains on changing task streams.
\end{compactenum}


\section{Related Work}
\label{sec:related}

\noindent\textbf{Continual Learning in Task Streams.}
Continual learning studies systems that learn from a sequence of tasks while retaining earlier capabilities~\cite{buzzega2020dark,wang2022learning,wang2022dualprompt}. Domain and test-time adaptation address distribution shift~\cite{ben-david2010da,ganin2016dann,wang2021tent,liang2020shot}, and mixture-of-experts methods route heterogeneous inputs to specialized components~\cite{jacobs1991moe,shazeer2017moe}. These directions cover pieces of our D1--D3 setting, but they usually adapt model weights, classifiers, or expert modules. In contrast, we study the harness-level analog.

\noindent\textbf{Self-improving and self-evolving agents.}
The closest LLM-agent precedents are systems that update harnesses from execution feedback. A-Evolve~\cite{lin2026position} uses a linear-chain evolver that mutates prompts, skills, memory, and tools. GEPA~\cite{agrawal2025gepa} adds reflective Pareto prompt evolution using textual feedback. Meta-Harness~\cite{lee2026meta} uses a growing filesystem archive and Claude Code as the proposer. Continual Harness~\cite{karten2026continual} supports online adaptation through alternating action/refinement cycles. SkillOS~\cite{ouyang2026skillos} trains a skill curator through reinforcement learning. OctoTools~\cite{lu2025octotools} supplies a human-designed tool-based agent framework. We adapt these approaches' evolution components to shared task-stream interfaces: GEPA and Meta-Harness selection is simplified, SkillOS uses heuristic curation, and the OctoTools toolkit is frozen (Appendix~\ref{app:baseline_protocol}). Our contribution is a framework for implementing and comparing ways to retain, revise, and deploy harness knowledge, together with a stateful multi-agent configuration and empirical deployment analyses. The comparisons concern these configurations, with the source methods' full native designs distinguished from their ports.

Appendix~\ref{app:related_work} provides the extended discussion.


\section{Adaptive Auto-Harness Framework}
\label{sec:method}

\begin{figure*}[t]
\centering
\includegraphics[width=0.95\textwidth]{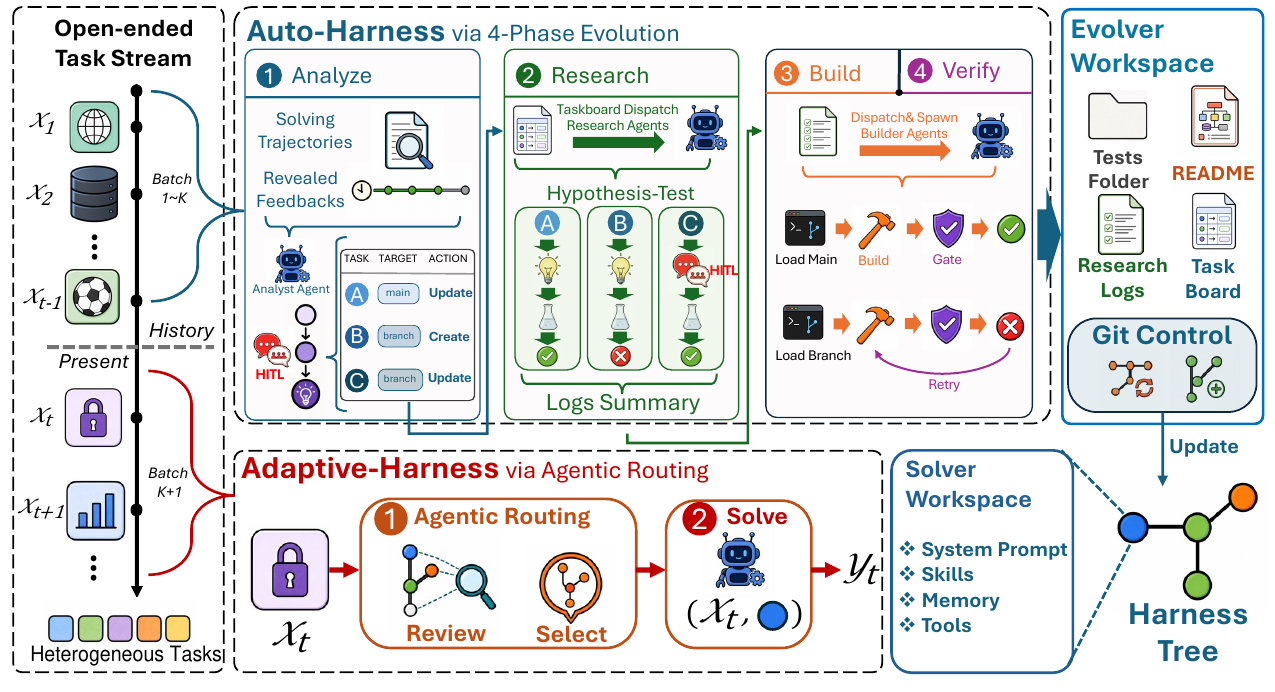}
\caption{Adaptive Auto-Harness with the stateful multi-agent evolution configuration. Analyze $\to$ Research $\to$ Build $\to$ Verify maintains research state and deployable harness versions. A temporal-reveal gate schedules explicit feedback; historical information boundaries are detailed in Appendix~\ref{app:benchmarks}. Deployment can use the current harness or a router that selects a branch for $x_t$. Optional human-in-the-loop (HITL) hooks on the task board and research phase accept operator guidance or access.}
\label{fig:unified_pipeline}
\end{figure*}

Adaptive Auto-Harness separates a common task-stream execution layer from
the procedure that updates the harness. The same execution interface hosts
different evolution strategies; each determines how to inspect experience,
maintain state, and revise deployable artifacts. Figure~\ref{fig:unified_pipeline}
illustrates our stateful multi-agent configuration and its deployment options.

\subsection{Problem Formulation}
\label{sec:method-setup}

We consider a finite prefix $x_1, x_2, \ldots, x_T$ of an open-ended task stream, with $x_t \sim P_t$, where $P_t$ is the task distribution at time $t$ and each $x_t$ has a fixed ground truth $y(x_t)$. The agent observes the history of prior experience $\Hist = \{(x_i, r_i, \tau_i)\}_{i=1}^{t-1}$, where $r_i$ is the realized reward once revealed and is otherwise unavailable, and $\tau_i$ is the observed solving trajectory (actions, intermediate observations, tool calls). A \emph{harness} $C_t = \varphi(\Hist)$, with $|C_t| \leq K$, is a bounded representation (prompts, skills, memory, and tools) with capacity budget $K$, where $\varphi$ is the \emph{evolver}: the agentic system that automatically transforms historic experience into the harness. A solver agent then acts via the policy $a_t \sim \pi(a \mid x_t, C_t)$, where $\pi$ is the LLM-induced sampling distribution over actions conditioned on the task and harness.

\subsection{Construction and Adaptation}
\label{sec:framework}

The three deployment dimensions of \S\ref{sec:intro} motivate two complementary design choices: how to construct and revise a harness from growing experience, and how to select a suitable harness for the incoming task. We organize these choices with a regret decomposition against an oracle reference. The identity holds at a fixed history and task distribution; experiments measure system contrasts and potential gains from harness selection, not the two oracle losses.

\noindent\textbf{Utility and regret.}
We define the utility of a harness $C$ on a task $x_t$ as
\begin{equation}
\label{eq:utility}
V(C, x_t) = \mathbb{E}_{a_t \sim \pi(\cdot \mid x_t, C)}[r(a_t, y(x_t))],
\end{equation}
which is the expected reward of the harness-conditioned solver. Let $\mathcal{C}_K(\Hist)$ denote the admissible harnesses under capacity budget $K$ and the available history and tool-access contract, without access to the unrevealed label $y(x_t)$. Here $K$ bounds representation capacity in the oracle comparison; it is not a matched evolver-compute budget. Define the conceptual full-history ceiling as
$V^\star(\Hist, x_t) \,:=\, \sup_{C\in\mathcal{C}_K(\Hist)} V(C, x_t)$.
The regret of an admissible evolver-constructed harness is then
\begin{equation}
\label{eq:regret}
\text{Regret}(\varphi, x_t) = V^\star(\Hist, x_t) - V(\varphi(\Hist), x_t) \geq 0,
\end{equation}
non-negative since $\varphi(\Hist)$ belongs to the supremum's domain. This reference does not imply that a deployed full-history solver can attain the ceiling under a finite computation budget.

\noindent\textbf{Solve-time optimal harness.}
Fix the evolver class $\Phi$ and define the solve-time optimal harness for a particular task:
\begin{equation}
\label{eq:cstar}
C^*_\Phi(x_t) = \arg\max_{\substack{C = \varphi(\Hist, x_t),\ \varphi \in \Phi \\ C\in\mathcal{C}_K(\Hist)}} V(C, x_t).
\end{equation}
This is a hypothetical oracle reference, assuming a maximizer exists: it is the best harness the class $\Phi$ could produce when allowed to condition on $x_t$. We assume that $\Phi$ also contains the deployed task-independent construction, represented by ignoring $x_t$. Adding and subtracting the pivot utility gives the following identity; the admissibility and inclusion assumptions make both terms non-negative.

\begin{proposition}[Regret Decomposition]
\label{prop:decomp}
For any deployed evolver $\varphi \in \Phi$, writing $C=\varphi(\Hist)$,
\begin{equation}
\label{eq:decomp}
\mathbb{E}_{x_t}[\text{Regret}(\varphi, x_t)] = \Levo(\Phi) + \Ladapt(\varphi),
\end{equation}
where
\begin{align}
\label{eq:losses}
\Levo(\Phi) &= \mathbb{E}_{x_t}\!\bigl[V^\star(\Hist, x_t) - V(C^*_\Phi(x_t), x_t)\bigr],
\end{align}
\begin{align}
\Ladapt(\varphi)  &= \mathbb{E}_{x_t}\!\bigl[V(C^*_\Phi(x_t), x_t) - V(C, x_t)\bigr].
\end{align}
\end{proposition}

$\Levo$ is the \emph{evolution loss}: the gap between the admissible ceiling and the best task-conditioned construction within $\Phi$. Restricting an evolver to prompt edits, for example, excludes tool implementations from its action space. A single agent with file-editing tools need not have this restriction. $\Ladapt$ is the \emph{adaptation loss}: the gap between the best task-conditioned construction within the class and the particular task-independent harness deployed. It can be positive when tasks prefer different harnesses, but heterogeneity alone does not guarantee a positive gap. Neither term identifies which architectural component caused a measured performance difference.


\subsection{$L_{evo}$: Harness Evolution}
\label{sec:method-multi}

The construction axis specifies how accumulated experience becomes a revised
harness. Evolution procedures share an execution interface and workspace
contract, while choosing their own history access, edit scope, and update
organization. Section~\ref{sec:setup} instantiates procedures adapted from prior
auto-harness work. Our stateful multi-agent configuration
(Figure~\ref{fig:unified_pipeline}) provides separate contexts and persistent
records for global-signal research and consolidation:

\noindent\textbf{(1) Multi-agent with distinct roles and objectives.}
We decompose evolution into four phases (Analyze $\to$ Research $\to$ Build $\to$ Verify), each with a dedicated objective and context. The Analyst prioritizes recurring failures; parallel Researchers investigate candidate explanations; Builders translate findings into harness changes; and Verifiers check the resulting artifacts with available tests.

\noindent\textbf{(2) Temporal-reveal feedback.}
Labels may arrive asynchronously, for example after a prediction market resolves. A \emph{temporal-reveal gate} is designed to delay explicit success/score feedback until resolution. This is a channel-level control, not an information-flow guarantee: answer-bearing archive metadata and historical retrieval require separate checks (Appendix~\ref{app:benchmarks}).

\noindent\textbf{(3) Persistent cross-cycle state.}
We provide the evolver with a dedicated workspace that persists across cycles, containing a task board (prioritized failure analysis), research logs (tested hypotheses with pass/fail verdicts), architecture documentation (README), and verification tests. This research workspace is separate from the solver's deployable prompts, skills, memory, and tools. Prior analyses remain available for later updates while the solver executes the current harness.

\subsection{$L_{adapt}$: Harness Storage and Deployment}
\label{sec:method-nav}

Deployment combines a storage structure with a selection operator:
a current harness, a tree of task-specific branches, or a skill catalog
with retrieval~\cite{deerflow2025}. Separating these choices lets the
framework study how the available knowledge and its selection affect reuse.

Our \emph{harness tree} stores prompts, skills, and tools in git branches
alongside a shared \texttt{main} branch, preserving versions and lineage.
When enabled, an LLM router inspects branch contents via \texttt{git show},
selects a branch for $x_t$, and the solver executes from that checkout.
Selection identifies the saved state; loading and tool invocation determine
its actual use. Main experiments compare complete configurations;
\S\ref{sec:navloss} examines selection within harness trees.
Branch initialization and diagnostic protocols are in
Appendix~\ref{app:branch_perf}.

\subsection{Optional Human Input}
\label{sec:method-hitl}

Optional task-board and research-assistance hooks accept operator guidance
or access (Figure~\ref{fig:unified_pipeline}). An operator may adjust task-board
priorities after analysis or provide access during research. The author-scripted
experiment in \S\ref{sec:hitl} documents messages and credentials; its protocol
and interpretation are detailed in Appendix~\ref{app:hitl_log}.


\section{Experiments}
\label{sec:experiments}

\subsection{Experimental Setup}
\label{sec:setup}

\noindent\textbf{Task streams.}
We use PolyBench prediction markets~\cite{cheng2026polybench},
CTF-Dojo security challenges~\cite{zhuo2025training}, and FutureX event
forecasts~\cite{zeng2025futurex} (Table~\ref{tab:bench_stats}), following
their recorded temporal orderings. They expose finite histories,
heterogeneous tasks, and changing input distributions.

\begin{table}[!t]
\centering
\footnotesize
\setlength{\tabcolsep}{2pt}
\caption{Benchmark statistics for the three chronological task streams.}
\vskip -1em
\label{tab:bench_stats}
\resizebox{\columnwidth}{!}{%
\begin{tabular}{@{}lrlll@{}}
\toprule
Bench. & Tasks & Span &Domain \\
\midrule
PolyBench & 5{,}075 & Feb 6--22, 2026 & Prediction markets \\
CTF-Dojo  & 261   & 2011--2024  & Security \\
FutureX   & 503   & Jan--Apr 2026 & Forecasting \\
\bottomrule
\end{tabular}
}
\vskip -1em
\end{table}

\providecommand{\vv}[1]{\makebox[2.45em][r]{#1}}
\providecommand{\nv}[1]{\makebox[3.0em][r]{#1}}
\providecommand{\cf}[1]{\raisebox{-0.35ex}{\scriptsize\,#1}}
\providecommand{\ci}[1]{\makebox[3.9em][l]{\raisebox{-0.35ex}{\scriptsize\,#1}}}
\begin{table*}[t]
\centering\footnotesize
\setlength{\tabcolsep}{4pt}
\caption{\zw{PolyBench results; subscripts are bootstrap 95\% CIs. \textbf{Cov} is \emph{coverage}, the share of markets traded, and \textbf{Acc} the share of \emph{all} served markets traded correctly, so declining to trade lowers Acc; \textbf{CV} is the \emph{coefficient of variation} of per-batch accuracy, i.e.\ stability across the stream; \textbf{Net\$} is realized money from a \$5{,}000 bankroll. Band columns bin markets by bought-side entry price: deepdog $<$0.3, dog 0.3--0.5, toss 0.5--0.7, fav 0.7--0.9, lock $\ge$0.9. Every system is near-perfect on \emph{lock} markets, so the score is decided in \emph{deepdog} and \emph{dog}, where a system must commit against the crowd. Market-following trades everything at 72\% accuracy yet loses \$5{,}348.}}
\label{tab:polybench_main}
\begin{tabular}{@{}llccccccccc@{}}
\toprule
& & & \multicolumn{5}{c}{Accuracy by entry-price band} & & & \\
\cmidrule(lr){4-8}
& System & Cov & deepdog & dog & toss & fav & lock & Acc & CV & Net\$ \\
\midrule
\multirow{5}{*}{\rotatebox[origin=c]{90}{\scriptsize No-evolution}}
& Static Sonnet & \vv{31.5}\ci{[30.2,32.7]} & 29 & 62 & 71 & 88 & \phantom{0}98 & \vv{22.2}\ci{[21.0,23.3]} & 0.26 & \nv{$+403$}\cf{[$-11$,$+825$]} \\
& Haiku & \vv{32.6}\ci{[31.3,33.9]} & \phantom{0}9 & 46 & 57 & 86 & \phantom{0}97 & \vv{15.2}\ci{[14.2,16.2]} & 0.48 & \nv{$-1214$}\cf{[$-1743$,$-630$]} \\
& DeepSeek & \vv{\phantom{0}2.9}\ci{[\phantom{0}2.4,\phantom{0}3.4]} & \phantom{0}9 & 46 & 63 & 88 & 100 & \vv{\phantom{0}1.4}\ci{[\phantom{0}1.1,\phantom{0}1.8]} & 1.03 & \nv{$-159$}\cf{[$-265$,$-50$]} \\
& Kimi & \vv{24.7}\ci{[23.6,25.8]} & 14 & 49 & 62 & 86 & \phantom{0}98 & \vv{14.0}\ci{[13.1,14.9]} & 0.46 & \nv{$-343$}\cf{[$-805$,$+116$]} \\
& GLM & \vv{21.8}\ci{[20.7,22.8]} & 19 & 55 & 66 & 84 & \phantom{0}97 & \vv{14.0}\ci{[13.0,14.8]} & 0.48 & \nv{$-25$}\cf{[$-398$,$+387$]} \\
\midrule
\multirow{5}{*}{\rotatebox[origin=c]{90}{\scriptsize Single-agent}}
& A-Evolve & \vv{21.1}\ci{[20.1,22.3]} & 69 & 82 & 89 & 96 & 100 & \vv{18.4}\ci{[17.4,19.4]} & 0.68 & \nv{$+684$}\cf{[$+426$,$+974$]} \\
& GEPA & \vv{22.6}\ci{[21.5,23.7]} & 22 & 48 & 61 & 87 & \phantom{0}97 & \vv{13.4}\ci{[12.5,14.4]} & 0.46 & \nv{$+269$}\cf{[$-115$,$+675$]} \\
& Meta-Harness & \vv{55.3}\ci{[53.9,56.7]} & 74 & 84 & 87 & 87 & \phantom{0}99 & \vv{50.8}\ci{[49.4,52.2]} & 0.42 & \nv{$+798$}\cf{[$+524$,$+1093$]} \\
& Continual & \vv{93.3}\ci{[92.6,94.0]} & 59 & 76 & 70 & 86 & \phantom{0}99 & \vv{76.9}\ci{[75.8,78.1]} & 0.11 & \nv{$+693$}\cf{[$+384$,$+1012$]} \\
& SkillOS & \vv{24.1}\ci{[23.1,25.2]} & 41 & 80 & 84 & 89 & \phantom{0}99 & \vv{21.4}\ci{[20.4,22.5]} & 1.37 & \nv{$+720$}\cf{[$+392$,$+1061$]} \\
\midrule
{\scriptsize Human} & OctoTools & \vv{55.6}\ci{[54.3,56.9]} & 20 & 69 & 74 & 88 & \phantom{0}99 & \vv{40.0}\ci{[38.7,41.4]} & 0.29 & \nv{$-368$}\cf{[$-869$,$+104$]} \\
\midrule
{\scriptsize Trivial} & Market-following & \vv{100.0}\ci{[100,100]} & --- & --- & 52 & 82 & \phantom{0}99 & \vv{72.0}\ci{[70.8,73.1]} & 0.22 & \nv{$\mathbf{-5348}$}\cf{[$-6.3$k,$-4.4$k]} \\
\midrule
\multirow{3}{*}{\rotatebox[origin=c]{90}{\scriptsize \textbf{Ours}}}
& \textbf{Multi-agent} & \vv{95.8}\ci{[95.3,96.4]} & 80 & 76 & 72 & 86 & \phantom{0}99 & \vv{79.8}\ci{[78.8,81.0]} & \textbf{0.10} & \nv{$+793$}\cf{[$+497$,$+1091$]} \\
& \textbf{Adaptive} & \vv{91.4}\ci{[90.7,92.2]} & 71 & 81 & 72 & 86 & \phantom{0}99 & \vv{77.4}\ci{[76.3,78.6]} & 0.15 & \nv{$\mathbf{+949}$}\cf{[$+689$,$+1221$]} \\
& \textbf{Full System} & \vv{\textbf{97.9}}\ci{[97.5,98.3]} & \textbf{92} & 80 & 68 & 86 & \phantom{0}99 & \vv{\textbf{80.9}}\ci{[79.8,82.0]} & 0.11 & \nv{$+881$}\cf{[$+511$,$+1261$]} \\
\bottomrule
\end{tabular}
\end{table*}

\begin{table}[t]
\centering\footnotesize
\setlength{\tabcolsep}{2pt}
\caption{CTF-Dojo and FutureX Pass@1 (\%), with 95\% task-bootstrap CIs. FutureX is exploratory.}
\label{tab:ctf_futurex_main}
\begin{tabular}{@{}lcc@{}}
\toprule
System & CTF-Dojo & FutureX \\
\midrule
Static Sonnet & 37.2\,{\scriptsize [31.4, 42.9]} & 31.0\,{\scriptsize [27.0, 35.0]} \\
A-Evolve & 45.2\,{\scriptsize [39.1, 51.3]} & 47.5\,{\scriptsize [43.1, 51.9]} \\
GEPA & 42.9\,{\scriptsize [36.8, 49.0]} & 28.2\,{\scriptsize [24.5, 32.2]} \\
Meta-Harness & 41.0\,{\scriptsize [35.2, 47.1]} & 29.4\,{\scriptsize [25.4, 33.4]} \\
Continual & 46.4\,{\scriptsize [40.2, 52.5]} & 31.0\,{\scriptsize [27.0, 35.0]} \\
SkillOS & 29.5\,{\scriptsize [24.1, 35.2]} & 29.8\,{\scriptsize [25.8, 33.8]} \\
OctoTools & 36.0\,{\scriptsize [30.3, 41.8]} & 43.3\,{\scriptsize [39.0, 47.7]} \\
\midrule
Adaptive & 46.0\,{\scriptsize [39.8, 52.1]} & 44.1\,{\scriptsize [40.0, 48.5]} \\
Multi-agent & 47.9\,{\scriptsize [41.8, 53.6]} & 49.5\,{\scriptsize [45.1, 53.9]} \\
Full System & 50.2\,{\scriptsize [44.1, 56.3]} & 47.3\,{\scriptsize [42.9, 51.7]} \\
\bottomrule
\end{tabular}
\end{table}

\noindent\textbf{Configurations and controls.}
The shared framework hosts update procedures adapted from A-Evolve,
GEPA, Meta-Harness, Continual Harness, and SkillOS, plus Adaptive,
Multi-agent, and our full system (Full System in tables; Ours in
Figure~\ref{fig:q1_dynamics}). Comparisons use Sonnet~4.6
for solving and Opus~4.6 for evolution at temperature~0.
Controls include static Sonnet, four other fixed solvers, frozen OctoTools,
and a market-following policy. These are implemented evolution variants,
with shared task sets and solver interfaces but unmatched total evolver
compute. Detailed configurations, source citations, nominal schedules, and
resource records are in Appendices~\ref{app:baseline_protocol}--\ref{app:experimental-details}.

\noindent\textbf{Evaluation.}
PolyBench separates participation (coverage), all-task accuracy, and
conditional accuracy; CTF-Dojo and FutureX use Pass@1.
Statistical comparisons and uncertainty estimates are reported in
Table~\ref{tab:headline_pairs}. FutureX is exploratory because premature
answer exposure and inconsistent retrieval cutoffs remain unresolved.
Benchmark contracts and information boundaries are in
Appendix~\ref{app:benchmarks}.

\noindent\textbf{Research questions.}
\textbf{RQ1} (\S\ref{sec:main-results}): How do evolution strategies compare within a shared task-stream framework?
\textbf{RQ2} (\S\ref{sec:ablation}): What drives performance gains from harness evolution?
\textbf{RQ3} (\S\ref{sec:revision}): Does harness evolution keep turning experience into gains as task streams grow, diversify, and shift?
\textbf{RQ4} (\S\ref{sec:hitl}): How do author-written steering messages compare in an access-limited experiment?

\subsection{Performance on Task Streams}
\label{sec:main-results}

\noindent\textbf{Strong gains over a fixed solver.}
With Sonnet fixed and no human input, our full system reaches $80.9\%$ PolyBench
accuracy versus $22.2\%$ without evolution, and $50.2\%$ CTF-Dojo Pass@1
versus $37.2\%$
(Tables~\ref{tab:polybench_main}--\ref{tab:ctf_futurex_main}).
Our full system exceeds Continual on PolyBench by $4.0$ points; differences among
the leading evolved configurations are smaller than their gains over
static Sonnet
(Table~\ref{tab:headline_pairs}).

\begin{figure}[!t]
\centering
\includegraphics[width=\columnwidth]{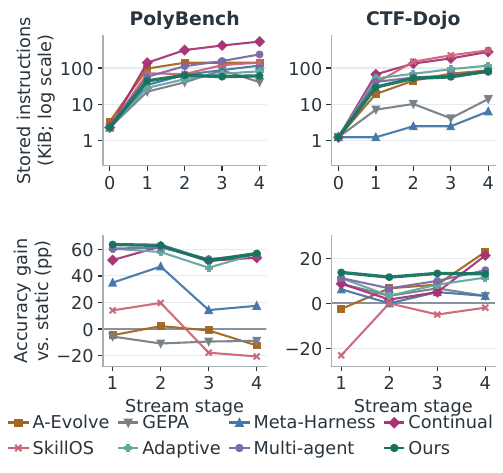}
\caption{Stored instructions (top) and same-task gains over static Sonnet (bottom). Lines denote configurations; stage~0 is the initial harness.}
\label{fig:q1_dynamics}
\end{figure}

\noindent\textbf{Artifact growth and performance can diverge.}
Instruction volume increases from the initial to the final harness
in all 16 runs (Figure~\ref{fig:q1_dynamics}), yet four runs underperform
static Sonnet overall.
Our full system retains same-task gains in every stage of both streams.
A-Evolve's CTF-Dojo margin rises from $-2.5$ to $+23.0$ points.
SkillOS's PolyBench gain instead reverses from $+19.6$ to $-20.7$ points
during stages~2--4, as its instruction volume grows $107.8\%$ and its answering rate
falls from $47.2\%$ to $3.5\%$.
Input shifts and task-level decompositions are in
Appendix~\ref{app:nonstationarity}.

\noindent\textbf{Participation and decision quality are distinct.}
Our full system and Continual have similar conditional accuracy ($82.6/82.4\%$)
at different coverage ($97.9/93.3\%$); their all-task accuracy gap mainly
reflects participation. Yet our full system exceeds the always-participating market
control by $8.9$ points ($80.9/72.0\%$).
This rules out simply following market consensus as the explanation,
while leaving comparison with an always-answer static LLM open.

\noindent\underline{\textbf{Takeaway~1.}}
As harness artifacts accumulate, gains over a fixed solver can
persist, emerge late, or reverse. Evaluating experience reuse therefore
requires tracking the performance margin across the stream; aggregate
accuracy and artifact growth can hide the loss of earlier gains.

\subsection{Drivers of Performance Gains}
\label{sec:ablation}

\begin{figure}[!t]
\centering
\includegraphics[width=\linewidth]{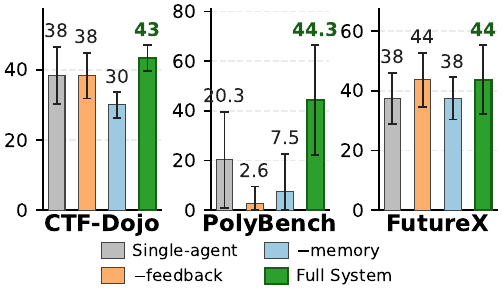}
\vskip -0.5em
\caption{Our full system versus intended ablations of multi-agent evolution, feedback, and persistent memory. PolyBench uses confidence-weighted return (CWR); the others use Pass@1.}
\label{fig:rq2_stats}
\vskip -0.5em
\end{figure}

\noindent\textbf{The memory comparison shows a consistent gap.}
The variant intended to remove persistent memory has one of the largest
deficits relative to our full system in every stream (Figure~\ref{fig:rq2_stats}).
This pattern highlights the potential importance of persistent research
state for accumulating experience across evolution cycles and informing
subsequent harness updates.

\begin{figure}[!t]
\centering
\includegraphics[width=0.9\linewidth]{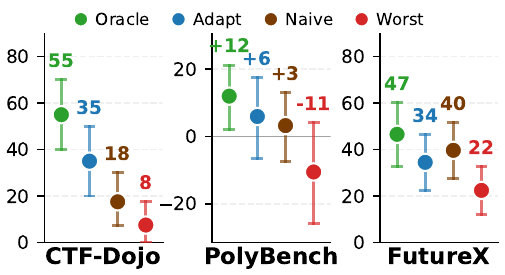}
\vskip -0.5em
\caption{Routing within harness trees: best/worst hindsight replay (Oracle/Worst), fixed-branch replay (Naive), and original router outcomes (Adapt).}
\label{fig:rq3_stats}
\vskip -0.5em
\end{figure}

\noindent\textbf{Better alternatives do not ensure effective selection.}
\label{sec:navloss}
In the separate harness trees of Figure~\ref{fig:rq3_stats},
Oracle exceeds Naive by $37.5$ points on CTF-Dojo
but only $6.9$ on FutureX. Adapt leads Naive on CTF-Dojo and PolyBench,
yet falls $5.2$ points below it on FutureX despite the higher Oracle score
(Appendix~\ref{app:branch_perf}; Table~\ref{tab:routing_stats}).

\begin{figure}[!t]
\centering
\includegraphics[width=\linewidth]{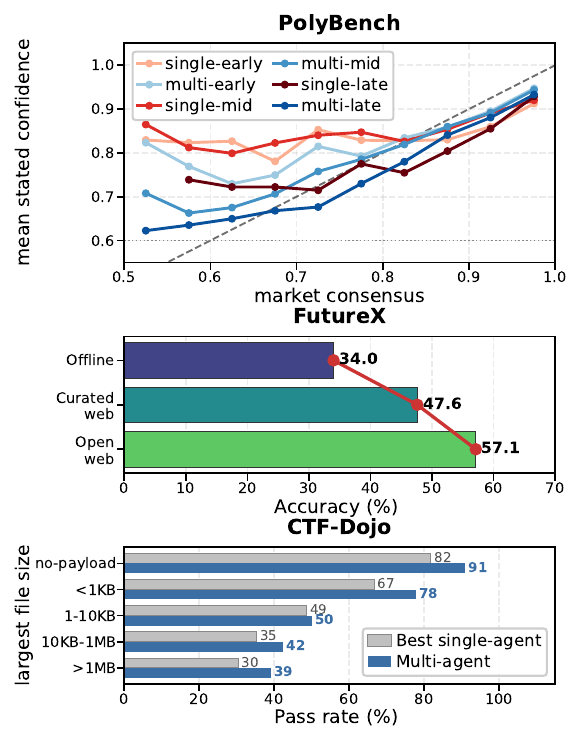}
\vskip -1em
\caption{\textbf{Capability bottlenecks for harness evolution.} Confidence becomes more sensitive to market consensus (PolyBench); broader web access accompanies higher accuracy (FutureX); large payloads remain difficult despite multi-agent gains (CTF-Dojo).}
\label{fig:l_evo_evidence}
\vskip -0.5em
\end{figure}

\noindent\textbf{Capability bottlenecks persist under stronger evolution.}
In Figure~\ref{fig:l_evo_evidence}, late multi-agent confidence on PolyBench
falls most on weak-consensus markets while remaining high when consensus
is strong.
FutureX's offline, curated-web, and open-web groups reach
$34.0/47.6/57.1\%$ accuracy, respectively.
On CTF-Dojo, the multi-agent configuration scores higher in every
payload-size group, yet its Pass@1 falls from $91\%$ without payloads
to $39\%$ above 1\,MB.
These patterns identify distinct construction priorities for reducing
$\Levo$: \emph{confidence calibration}, \emph{evidence acquisition},
and \emph{execution on large inputs}
(Appendix~\ref{app:capability-diagnostics}).

\begin{figure*}[!t]
\centering
\includegraphics[width=\textwidth]{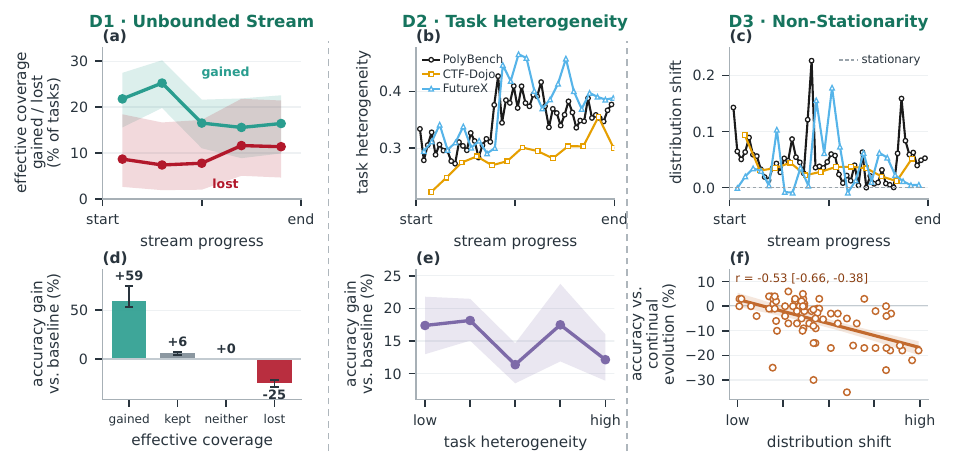}
\caption{\textbf{(a)~Effective coverage gains do not grow.} Tasks whose effective coverage is gained or lost against static Sonnet. \textbf{(b)~Heterogeneity rises.} Mean distance to all earlier tasks. \textbf{(c)~Shifts occur throughout.} Energy distance to the batches the continual harness last learned from (colors as in~b). \textbf{(d)~Gain follows effective coverage.} Gain per effective-coverage transition. \textbf{(e)~Heterogeneity: no gain trend.} Gain by heterogeneity quintile. \textbf{(f)~Shift hurts frozen harnesses.} Frozen minus continual accuracy against distance from the freeze point. Details: Appendix~\ref{app:stream-dimensions}.}
\label{fig:dimension_correlations}
\end{figure*}

\noindent\underline{\textbf{Takeaway~2.}}
Stronger harness evolution can raise overall performance while leaving
capability bottlenecks unresolved. Sustained gains require turning
accumulated experience into changes to the operations that still limit
the solver on incoming tasks.

\subsection{Sustained Gains as Task Streams Change}
\label{sec:revision}
\label{sec:barriers}

\noindent\textbf{Effective coverage gains do not keep expanding.}
Figure~\ref{fig:dimension_correlations} first relates gains over static
Sonnet to \emph{effective coverage}: committing a trade, submitting a
non-placeholder flag, or using a tool before forecasting.
Gains average $+59.1$ points when only the evolved solver achieves effective
coverage, $-24.8$ when only static Sonnet does, and $+5.7$ when both do
(panel~d). Yet gained effective coverage peaks early and ends at
$16.4\%$ of tasks, while lost effective coverage rises from $8.7\%$ to
$11.4\%$ (panel~a).

\noindent\textbf{Greater heterogeneity need not reduce gains.}
\emph{Task heterogeneity}, measured by mean distance to all earlier tasks
over solver-visible properties, increases as each stream grows (panel~b).
Yet its association with gains has opposite signs across streams,
yielding a near-zero pooled correlation ($r=-0.06$; panel~e).

\noindent\textbf{Continued updates preserve gains under shift.}
Task heterogeneity alone does not capture changes since the last harness
update. We measure \emph{distribution shift} by energy distance between
incoming batches and the tasks preceding that update
(panel~c). Replays on two streams compare frozen and continually updated
harnesses on the same later tasks: frozen harnesses fall further behind
as shift grows ($r=-0.53$; panel~f).

\noindent\underline{\textbf{Takeaway~3.}}
Knowledge that transfers across diverse tasks can still lose value as
the task distribution changes. Sustained improvement can therefore require
continued revision of accumulated knowledge to preserve earlier gains.

\subsection{\zw{Human-in-the-Loop Steering}}
\label{sec:hitl}

\begin{figure}[!t]
\centering
\includegraphics[width=\linewidth]{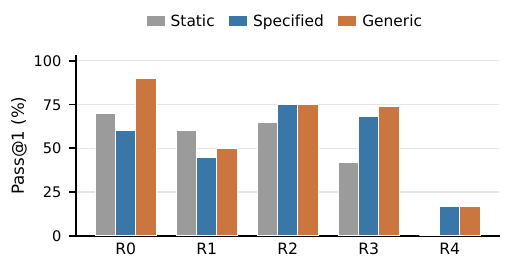}
\caption{Steering by retrieval tier (R0--R4). Specified and Generic evolve with search credentials; Static is frozen and keyless. Protocol: Appendix~\ref{app:hitl_log}.}
\label{fig:rq4_hitl_passrate}
\vskip -0.5em
\end{figure}

\noindent\textbf{Specific guidance provides no observed advantage.}
In the 85-task FutureX experiment, author-written messages either recommend
particular types of primary sources (\emph{Specified}) or ask the evolver
to improve retrieval without prescribing sources (\emph{Generic};
Figure~\ref{fig:rq4_hitl_passrate}).
The messages arrive after the first 20 tasks; both steered systems evolve
with search credentials. On the same 65 subsequent tasks, Specified and
Generic achieve Pass@1 of $61.5\%$ and $73.8\%$, respectively,
versus $55.4\%$ for the static keyless reference
(Appendix~\ref{app:hitl_log}).

\noindent\underline{\textbf{Takeaway~4.}}
More specific instructions do not guarantee more useful guidance.
The steered systems also receive search credentials and evolve, so
their gains over the static solver cannot be attributed to guidance alone.


\section{Conclusions}
\label{sec:conclusions}
\vspace{-0.35em}

On open-ended task streams, successful experience reuse depends on whether
an agent's advantage persists as incoming tasks change. We introduce Adaptive
Auto-Harness to study this problem through a shared framework for evolution
and deployment. Recorded comparisons demonstrate gains with fixed solver
weights, while longitudinal analyses show that growing artifact collections
can accompany persistent, delayed, or reversed gains.

These findings place the durability of improvement at the center of
task-stream deployment. Increasing task heterogeneity has no consistent
relationship with gains across streams, yet frozen harnesses fall further
behind continued evolution as distribution shift grows. Continued evolution
can therefore have value in preserving an existing advantage, even when
improvement is not monotonic. For ongoing services, accumulated experience
is useful to the extent that it continues to improve subsequent decisions
and actions. Evaluation should track that advantage over a fixed solver
on the same incoming tasks, alongside changes in artifacts and solver
behavior throughout deployment.

\clearpage

\section{Limitations}
\label{sec:limitations}

\noindent\textbf{Evaluation scope.}
Experiments observe finite prefixes of three task streams, with one recorded
evolution history per configuration. Task and batch-block intervals quantify
uncertainty conditional on those histories, not independent evolution runs.
Long-term artifact aging and cost growth remain untested. Closed APIs and
tool nondeterminism limit exact reruns; deterministic statistical recomputation
requires the identified records and dataset snapshots.

\noindent\textbf{Comparison and attribution.}
The ported evolution configurations differ in selection mechanisms, recorded
schedules, and total evolver resources. Complete construction and billing
costs are unavailable. Results therefore concern these implementations;
component effects remain unisolated where intended feedback or memory
interventions were unverified, or where routing and steering comparisons
also change histories or access.

\noindent\textbf{Historical information access.}
FutureX contains confirmed premature answer exposure and inconsistent
retrieval cutoffs; its scores remain exploratory, and excluding flagged
tasks cannot repair affected evolution histories. PolyBench uses frozen
inputs, but its resolution-ordered replay and answer-bearing trajectory
indices do not certify prospective information isolation; actual premature
reads and their score impact remain unresolved. Decision accuracy and
simulated payoffs do not establish achievable trading profit.

Further evaluation details appear in
Appendices~\ref{app:benchmarks}--\ref{app:experimental-details}.


\section{Ethics Statement}
\label{sec:ethics}

We use public research benchmark tasks and do not introduce private user data. CTF-Dojo runs only inside isolated benchmark containers and does not target real systems. Human steering is limited to source guidance, task-board edits, and credential decisions; humans do not label/expose answers or choose solver branches.


\section{AI Usage Statement}
\label{sec:ai-usage}

We used AI assistants to refine the writing of this paper and to accelerate debugging and analysis during implementation and evaluation.

\bibliography{references}

\appendix
\raggedbottom
\setcounter{topnumber}{4}
\setcounter{bottomnumber}{3}
\setcounter{totalnumber}{6}
\setcounter{dbltopnumber}{4}
\renewcommand{\topfraction}{0.95}
\renewcommand{\bottomfraction}{0.90}
\renewcommand{\textfraction}{0.05}
\renewcommand{\floatpagefraction}{0.65}
\renewcommand{\dbltopfraction}{0.95}
\renewcommand{\dblfloatpagefraction}{0.65}
\setlength{\floatsep}{8pt plus 2pt minus 2pt}
\setlength{\textfloatsep}{10pt plus 2pt minus 2pt}
\setlength{\intextsep}{8pt plus 2pt minus 2pt}
\setlength{\dblfloatsep}{8pt plus 2pt minus 2pt}
\setlength{\dbltextfloatsep}{10pt plus 2pt minus 2pt}
\makeatletter
\setlength{\@fptop}{0pt}
\setlength{\@fpsep}{10pt}
\setlength{\@fpbot}{0pt plus 1fil}
\setlength{\@dblfptop}{0pt}
\setlength{\@dblfpsep}{10pt}
\setlength{\@dblfpbot}{0pt plus 1fil}
\makeatother


\clearpage
\section{Extended Related Work}
\label{app:related_work}

\noindent\textbf{Continual learning and task-stream evaluation.}
Continual learning studies how systems acquire new knowledge while
retaining earlier capabilities~\cite{kirkpatrick2017ewc,buzzega2020dark,delange2021cl}.
Prompt-based methods learn task-dependent prompts for a frozen
backbone~\cite{wang2022learning,wang2022dualprompt}, while domain and
test-time adaptation address changing input
distributions~\cite{ben-david2010da,ganin2016dann,liang2020shot,wang2021tent}.
For language agents, LifelongAgentBench organizes skill-related tasks
in database, operating-system, and knowledge-graph environments into
sequences to evaluate knowledge
accumulation~\cite{zheng2025lifelongagentbench}.
LiveCodeBench and SWE-bench-Live update task collections as new problems
appear~\cite{jain2024livecodebench,zhang2025swebenchlive}.
We evaluate harness evolution on finite streams assembled from existing
benchmarks in prediction markets, security challenges, and event forecasting,
with the solver's model weights held fixed. The deployment problem combines
growing histories, heterogeneous tasks, and changing distributions.

\noindent\textbf{Experience, memory, and reusable knowledge.}
Reflexion stores verbal feedback for later
trials~\cite{shinn2023reflexion}, while ExpeL extracts insights from
collections of agent experiences~\cite{zhao2024expel}.
Voyager retains executable skills~\cite{wang2023voyager}, and Agent
Workflow Memory induces reusable workflows in both offline and online
settings, including tests of cross-task and cross-domain
generalization~\cite{wang2024awm}.
Generative Agents, MemoryBank, and A-Mem organize and update external
memories~\cite{park2023generative,zhong2024memorybank,xu2025amem}.
Dynamic Cheatsheet and ACE accumulate and curate knowledge in evolving
contexts~\cite{suzgun2025dynamic,zhang2025ace}; ACE also analyzes knowledge
loss under repeated context rewriting. LIMBO allocates memory and
inference budgets per incoming task~\cite{sharma2026limbo}.
Our design separates a persistent research workspace from the solver's
deployable harness. This organization supports revisiting earlier analyses
while revising the prompts, skills, and tools used on subsequent tasks.

\noindent\textbf{Optimizing prompts and agent workflows.}
OPRO, ProTeGi, and TextGrad use model-generated proposals or textual
feedback to optimize instructions and system
components~\cite{yang2023opro,pryzant2023automatic,yuksekgonul2024textgrad}.
DSPy optimizes programs composed of language-model
calls~\cite{khattab2023dspy}.
GEPA combines reflection with Pareto selection over candidate
prompts~\cite{agrawal2025gepa}.
Automated Design of Agentic Systems, AFlow, and the Darwin G\"odel
Machine extend search to agent code, workflows, and evolving agent
archives~\cite{hu2024adas,zhang2024aflow,zhang2025dgm}.
These approaches provide mechanisms for improving external agent state.
We study repeated harness construction together with task-conditioned
deployment as experience and task requirements change.

\noindent\textbf{Harness evolution and deployment frameworks.}
A-Evolve studies evolution of persistent agent
state~\cite{lin2026position}. Meta-Harness searches harness code using
retained candidates, scores, and execution traces; its applications include
online text classification~\cite{lee2026meta}.
Continual Harness alternates acting and refinement within a persistent
embodied episode and compares frozen and updating
bootstraps~\cite{karten2026continual}.
SkillOS formulates skill maintenance on task streams and trains a curator
using subsequent-task utility; it also analyzes changes in retained
skills and their actual use~\cite{ouyang2026skillos}.
Task streams and ongoing maintenance are therefore shared with close
prior work. Adaptive Auto-Harness combines persistent multi-agent research,
harness revision, and task-conditioned deployment to address the joint
deployment pressures described in \S\ref{sec:intro}.
Multi-agent orchestration and tool interfaces are established by
AutoGen, MetaGPT, and
OctoTools~\cite{wu2023autogen,hong2023metagpt,lu2025octotools};
task-conditioned selection also has precedents in mixture-of-experts
models~\cite{jacobs1991moe,shazeer2017moe}.
Our comparisons concern the streaming implementations documented in
Appendix~\ref{app:baseline_protocol}, whose departures from native methods
limit claims about the original systems.


\section{Benchmark and Evaluation Details}
\label{app:benchmarks}

All main-table systems use the same task-ID set within each benchmark.
We retain the first record per task ID and follow the
benchmark-specific orderings below. The reveal gate schedules explicit
feedback; recorded schedule exceptions appear in Table~\ref{tab:batch_exceptions}.

\subsection{PolyBench}

\noindent\textbf{Source and composition.}
PolyBench contains 5{,}075 Polymarket tasks from Feb 6--22, 2026.
Each input freezes the order book and news context; the solver has no
live-web access. Scoring uses the official market outcome.

\noindent\textbf{Temporal ordering and information access.}
The queue follows resolution time, then snapshot time; all 50 adjacent
batch boundaries for our full system overlap in snapshot time.
Snapshot-local news may describe an event before its market settles
(Figure~\ref{fig:c1}). Separately, all 5,075 trajectory-index rows in the
retained archive contain outcome-matched PASS/FAIL labels. Premature reads
and their effects are unresolved. These are retrospective decision results;
freezing solver inputs alone does not establish prospective isolation.

\noindent\textbf{Metrics.}
Let $N$ count served tasks, $\mathcal{T}$ the ungated trade decisions,
and $s_i\in\{0,1\}$ directional correctness. Coverage concerns decisions,
independently of simulated fills:
\begin{align*}
\mathrm{Coverage} &= 100\cdot\tfrac{|\mathcal{T}|}{N}, \\
\mathrm{Acc}      &= 100 \cdot \tfrac{1}{N}\textstyle\sum_{i\in\mathcal{T}} s_i, \\
\mathrm{Acc}_{\mathrm{traded}} &= 100\cdot\tfrac{1}{|\mathcal{T}|}
\textstyle\sum_{i\in\mathcal{T}} s_i.
\end{align*}
Thus $\mathrm{Acc}=(\mathrm{Coverage}/100)\,\mathrm{Acc}_{\mathrm{traded}}$;
conditional accuracy compares system-selected subsets.
CWR is $100\sum_i g_i/\sum_i b_i$, with simulated investment $b_i$ and
profit $g_i$; coverage-scaled CWR additionally multiplies by $|\mathcal{T}|/N$.
These return ratios differ from all-task accuracy.

\noindent\textbf{Decision and fill rules.}
JSON parsing, with a free-form fallback, extracts the decision, outcome
side, and confidence. Gated or empty outputs, \textsc{skip}, and confidence
below $0.6$ remain in $N$ but are excluded from $\mathcal{T}$.
A buy is correct if its side wins; a sell is correct if it loses.
For monetary diagnostics, asks in $(0,1)$ are consumed in price order
up to $10\times\text{confidence}$ or available depth. Profit is
$\text{shares}-\text{spent}$ when correct and $-\text{spent}$ otherwise.
CWR can fall back to the snapshot mid-price (default $0.5$);
Table~\ref{tab:polybench_main}'s Net\$ has no price fallback.

\noindent\textbf{Net dollar payoff.}
Net\$ in Table~\ref{tab:polybench_main} is the simulated realized profit or
loss for a nominal \$5,000 starting bankroll, not the ending account balance.
It sums independently scored trades at resolution under the fill rules
above, with zero fees and no shared capital constraint.
A trade additionally requires a listed entry price
$\geq0.01$ and confidence minus price $\geq0.20$; these filters do not
define Coverage or Accuracy. Unfilled entries contribute no payoff.
These payoff diagnostics do not establish actual fills or achievable trading profit.

\noindent\textbf{Matched market-following control.}
The control buys the highest-price outcome on the same 5,075 inputs,
with confidence 1, listed-order tie breaking, and no search or LLM.
It makes 3,654 correct decisions (72.0\%, 100\% coverage), versus our
full system's 4,105 at 97.9\% coverage: $+8.9$ points
(batch-block CI $[5.6,13.8]$; Table~\ref{tab:headline_pairs}).
This advantage exceeds mechanical market-following despite lower
participation. An always-answer static LLM is a different, unevaluated
control.

\subsection{CTF-Dojo}

The 261 challenges from the pwn.college CTF archive follow the released
2011--2024 ordering and cover exploitation, web security, cryptography,
reverse engineering, and forensics. Each runs in a per-task Docker sandbox
under the benchmark network policy. Submitted flags are verified by
SHA-256 comparison with the official flag. Pass@1 is the percentage solved
within the task budget.

\subsection{FutureX}

\noindent\textbf{Source and composition.}
FutureX-Past supplies 503 English and Chinese questions over 82 days in
Jan--Apr 2026. We use its official Pass@1 criterion, including for HITL.

\noindent\textbf{Temporal retrieval and filtering.}
Safeguards include pre-cutoff Wikipedia revisions, publication-date filters,
bounded economic queries, and a separate LLM redaction call whose raw input
does not enter the solver context. Missing creation dates are estimated as
resolution minus seven days. General retrieval adds seven days to that
estimate; Wikipedia uses the estimate, and the benchmark helper allows a
one-day buffer. These cutoffs do not define a common historical snapshot.

\noindent\textbf{Exclusion sensitivity and scope.}
Removing the same 35 retrieval-flagged tasks from all systems leaves 468
tasks. Multi-agent, A-Evolve, and our full system score
$50.4/49.6/49.1\%$, compared with $49.5/47.5/47.3\%$ on all 503 tasks:
their ordering is unchanged. The retrieval audit covers only part of the
Multi-agent context. Separately, archived observations retain answer fields;
five unrevealed answers reached A-Evolve's evolution context, and
Multi-agent notes also quote unrevealed answers. Task exclusions cannot
remove information already incorporated into evolution, so these results
remain exploratory.

\section{Evolution Variants and Execution}
\label{app:baseline_protocol}

Table~\ref{tab:baseline_protocol} specifies each implemented evolution
variant's history access, persistent state, editable artifacts, and update
procedure. Source-method names credit the origins of these components;
the comparisons concern their implementations in the shared framework.

\noindent\textbf{Fixed-harness controls.}
Static Sonnet shares the primary solver backbone. Other fixed solvers are
DeepSeek-V3.2~\cite{deepseek2024v3},
Claude Haiku 4.5~\cite{anthropic2025haiku45}, GLM-4.7~\cite{glm2024chatglm},
and Kimi-K2.5~\cite{moonshot2025kimik2}. OctoTools~\cite{lu2025octotools}
provides a frozen toolkit through the common solver. The market-following
control is defined in Appendix~\ref{app:benchmarks}.

\noindent\textbf{Shared execution contract.}
All update procedures share the batch runner and observation interface.
A \emph{cycle} is one scheduled evolution step after a batch; an \emph{update} is a cycle that changes the harness.
Sonnet~4.6~\cite{anthropic2026sonnet46} and Opus~4.6 serve as solver/evolver at
temperature~0. Tables~\ref{tab:hyperparams}--\ref{tab:seed_inventory}
specify budgets, network permissions, schedules, and initial artifacts.
Errors and timeouts count as failures; first-record aggregation avoids
selecting the best retry. Task IDs and solver interfaces are shared;
update organization and total construction resources vary.

\noindent\textbf{Recorded schedules.}
Table~\ref{tab:batch_exceptions} distinguishes changed cadence from
incomplete accounting. FutureX Adaptive uses 40-task batches instead of
20. Other listed gaps concern observation/history records; their causes
remain unresolved. OctoTools is frozen, and PolyBench Continual's missing
history entry covers the final 75 tasks rather than an established earlier
serving update.

\noindent\textbf{Preserved mechanisms and substantive deviations.}
GEPA~\cite{agrawal2025gepa} retains reflection/mutation, adds reflection
memory, and omits Pareto selection and accept-if-better rollback.
Meta-Harness~\cite{lee2026meta} proposes one candidate rather than two and
omits frontier selection. Continual~\cite{karten2026continual} substitutes
a tools pass for its sub-agent pass. SkillOS~\cite{ouyang2026skillos}
uses an LLM curator without its trained RL policy or compactness reward.
A-Evolve~\cite{lin2026position} retains experience-driven file editing.
These choices define the compared variants; the shared interface does not
establish equivalence to each source method's native design.

\begin{table*}[htbp]
\centering\footnotesize
\setlength{\tabcolsep}{4pt}
\caption{\textbf{Evolution components in the shared framework.} Method names identify source approaches; rows specify retained procedures and adaptations. Native selection/rollback differences are detailed in Appendix~\ref{app:baseline_protocol}; recorded resources appear in Table~\ref{tab:cost}.}
\label{tab:baseline_protocol}
\begin{tabular}{@{}>{\raggedright\arraybackslash}p{1.5cm}>{\raggedright\arraybackslash}p{2.15cm}>{\raggedright\arraybackslash}p{2.55cm}>{\raggedright\arraybackslash}p{2.6cm}>{\raggedright\arraybackslash}p{2.15cm}>{\raggedright\arraybackslash}p{2.95cm}@{}}
\toprule
System & Evolver roles/cycle & Reads each cycle & Persistent state grown & Operators & Process \\
\midrule
A-Evolve & 1 & last-2-cycle window & prompts, skills, memory, tools, infra & add, update, delete (per item) & react to window $\to$ free-edit; commit each cycle \\
\addlinespace[2pt]
GEPA & 2 (reflect, mutate) & batch + own $k{=}3$ reflections & prompt (+notes) & prompt rewrite & reflect $\to$ rewrite; no native Pareto selection \\
\addlinespace[2pt]
Meta-Harness & 1 (proposer) & batch + archive (via bash) & candidate archive and active harness & archive append; harness edit & browse history $\to$ propose; no native frontier selection \\
\addlinespace[2pt]
Continual & 4 (one per component) & batch trajectories & component-specific state & add, update, delete (per component) & 4 independent CRUD passes over the batch \\
\addlinespace[2pt]
SkillOS & 1 (curator) & batch + registry + effectiveness log & external skill repo & add, update, delete (per skill) & heuristic effectiveness-based curation; no trained RL policy \\
\addlinespace[2pt]
OctoTools & 0 (frozen) & --- & fixed toolkit & none & human toolkit installed once, frozen \\
\midrule
\textbf{Ours} & 4 roles: analyst, researchers, builder, verifier & batch + persistent evolution state & harness and research records & edit, reconcile, test & diagnose $\to$ research $\to$ build $\to$ verify \\
\bottomrule
\end{tabular}
\end{table*}

\begin{table*}[htbp]
\centering\footnotesize
\setlength{\tabcolsep}{5pt}
\caption{Recorded schedule and accounting exceptions. History total sums \texttt{batch\_total}; Different labels counts task-to-batch mismatches against the static stream. These result files have no duplicate task IDs. Adaptive's 40-task batches agree across all records; other gaps have unresolved provenance.}
\label{tab:batch_exceptions}
\begin{tabular}{@{}llrrrrr@{}}
\toprule
Benchmark & System & Results & Observations & History total & Batches & Different labels \\
\midrule
PolyBench & Continual & 5,075 & 5,075 & 5,000 & 51 & 0 \\
PolyBench & OctoTools & 5,075 & 5,052 & 5,052 & 51 & 966 \\
CTF-Dojo & Continual & 261 & 242 & 242 & 13 & 39 \\
CTF-Dojo & SkillOS & 261 & 252 & 252 & 13 & 100 \\
FutureX & Adaptive & 503 & 503 & 503 & 13 & 483 \\
\bottomrule
\end{tabular}
\end{table*}

\section{Implementation and Reproducibility}
\label{app:experimental-details}

\noindent\textbf{Code and recorded artifacts.}
The linked repository provides the framework, benchmark loaders/scorers,
launch configurations, and initial harnesses. Retained run artifacts
include per-task predictions/evaluations, solver trajectories, batch
assignments, and evolution histories. Per-task results and trajectories
will be released upon acceptance. Main-table runs use no human input
and one recorded evolution history per configuration.

\noindent\textbf{Offline statistical reproduction.}
Archived predictions, batch assignments, and usage records support
recomputation of rates, paired comparisons, resource summaries, and
diagnostic subsets without model or retrieval calls. Paired comparisons
require identical task IDs; input hashes and duplicate counts identify the
retained records. Offline checks reproduced the saved counts, estimates,
and intervals exactly. The statistical procedures are specified below.

\noindent\textbf{Execution settings.}
Table~\ref{tab:hyperparams} gives nominal settings; actual exceptions are
in Table~\ref{tab:batch_exceptions}. Hosted APIs keep model weights fixed;
local compute runs orchestration and benchmark sandboxes.

\begin{table}[!htbp]
\centering\footnotesize
\setlength{\tabcolsep}{4pt}
\caption{Nominal settings. Common values span all three benchmarks. \emph{EGL}: Expected-Gain-from-Learning trigger. Scheduled batches are update opportunities; executed updates and schedule exceptions appear in Tables~\ref{tab:cost} and~\ref{tab:batch_exceptions}.}
\label{tab:hyperparams}
\resizebox{\columnwidth}{!}{%
\begin{tabular}{@{}lccc@{}}
\toprule
Hyperparameter & PolyBench & CTF-Dojo & FutureX \\
\midrule
\multicolumn{4}{@{}l}{\emph{Common models and limits}} \\
Solver / router & \multicolumn{3}{c}{Sonnet 4.6} \\
Evolver & \multicolumn{3}{c}{Opus 4.6} \\
Solver / evolver temperature & \multicolumn{3}{c}{0.0} \\
Solver max turns & \multicolumn{3}{c}{80} \\
Evolver token-budget parameter & \multicolumn{3}{c}{128k} \\
EGL threshold / window & \multicolumn{3}{c}{0.05 / 3} \\
Parallel researchers & \multicolumn{3}{c}{3} \\
Build/verify attempts (prompt) & \multicolumn{3}{c}{3} \\
Routing confidence threshold & \multicolumn{3}{c}{0.7} \\
\midrule
\multicolumn{4}{@{}l}{\emph{Stream \& schedule}} \\
Total tasks & 5{,}075 & 261 & 503 \\
Batch size & 100 & 20 & 20 \\
Scheduled batches & 51 & 14 & 26 \\
Solve workers & 24 & 8 & 10 \\
Sandbox network (both) & none & none & bridge \\
\bottomrule
\end{tabular}}
\end{table}

\noindent\textbf{Seed harness.}
Table~\ref{tab:seed_inventory} lists the initial prompts and scaffolding;
skills, tools, and memory begin empty. OctoTools receives its frozen toolkit.
Full evolver and router templates appear in Appendix~\ref{app:prompts}.

\begin{table}[htbp]
\centering\footnotesize
\setlength{\tabcolsep}{6pt}
\caption{Seed harness shipped to the evolver before any cycle runs. \emph{Skills} counts top-level skill directories; \emph{Memory} counts JSONL entries; \emph{Infra} indicates whether the seed includes an infrastructure directory.}
\label{tab:seed_inventory}
\resizebox{\columnwidth}{!}{%
\begin{tabular}{@{}lrrrrl@{}}
\toprule
Benchmark & Prompt LOC & Skills & Tools & Memory & Infra \\
\midrule
PolyBench & 27 & 0 & 0 & 0 & no \\
CTF-Dojo & 24 & 0 & 0 & 0 & no \\
FutureX & 114 & 0 & 0 & 0 & yes \\
\bottomrule
\end{tabular}}
\end{table}

\noindent\textbf{Resource accounting.}
Table~\ref{tab:cost} reports all ten Sonnet-based systems, with explicit
denominators for observed token/step fields. Elapsed time covers all tasks;
evolver hours sum recorded cycles. Steps are logger counts, not universal
API-call counts. Complete evolver-token and billing records are unavailable,
so recorded time does not establish equal total compute or dollar cost.

\begin{table*}[!t]
\centering\footnotesize
\setlength{\tabcolsep}{4pt}
\caption{Recorded resources for all ten Sonnet-based systems. Means use observed fields, with token/step denominators in the last column; missing usage is not zero. Time covers all tasks. Evo.\ h sums recorded cycle time; Updates counts \texttt{mutated=true} entries. Accounting definitions and unavailable cost fields are specified in Appendix~\ref{app:experimental-details}.}
\label{tab:cost}
\resizebox{\textwidth}{!}{%
\begin{tabular}{@{}lrrrrrrr@{}}
\toprule
System & Input tok & Output tok & Steps & s/task & Evo.\ h & Updates & $n_{\rm tok}/n_{\rm step}$ \\
\midrule
\multicolumn{8}{l}{\textit{PolyBench ($N=5,075$)}} \\
Static Sonnet & 6,948 & 988 & 0.0 & 18.2 & 0.0 & 0 & 5075/5075 \\
A-Evolve & 71,998 & 822 & 0.7 & 17.0 & 2.4 & 20 & 5075/5075 \\
GEPA & 43,379 & 1,519 & 1.2 & 30.4 & 4.9 & 51 & 5075/5075 \\
Meta-Harness & 48,273 & 1,898 & 1.4 & 35.2 & 5.3 & 35 & 5075/5075 \\
Continual & 62,244 & 2,301 & 1.6 & 39.9 & 11.5 & 50 & 5075/5075 \\
SkillOS & 25,682 & 1,836 & 1.7 & 32.1 & 3.5 & 51 & 5075/5075 \\
OctoTools & 56,926 & 4,910 & 5.7 & 78.6 & 0.0 & 0 & 5075/5075 \\
Adaptive & 44,440 & 1,995 & 1.8 & 43.7 & 5.3 & 39 & 5075/5075 \\
Multi-agent & 52,058 & 2,852 & 4.7 & 55.3 & 26.6 & 51 & 5075/5075 \\
Full System & 46,077 & 2,406 & 4.4 & 42.2 & 22.9 & 40 & 5061/5061 \\
\midrule
\multicolumn{8}{l}{\textit{CTF-Dojo ($N=261$)}} \\
Static Sonnet & 599,073 & 9,254 & 90.6 & 308.4 & 0.0 & 0 & 231/255 \\
A-Evolve & 426,392 & 6,589 & 17.1 & 156.7 & 1.5 & 14 & 261/261 \\
GEPA & 581,731 & 8,230 & 49.4 & 250.3 & 0.4 & 13 & 252/261 \\
Meta-Harness & 564,169 & 8,715 & 50.3 & 275.9 & 1.6 & 2 & 253/261 \\
Continual & 636,943 & 9,467 & 56.2 & 271.1 & 3.9 & 13 & 251/261 \\
SkillOS & 356,380 & 4,950 & 30.8 & 224.2 & 1.4 & 12 & 258/261 \\
OctoTools & 753,148 & 11,274 & 83.6 & 302.7 & 0.0 & 0 & 235/251 \\
Adaptive & 685,730 & 7,648 & 28.8 & 281.9 & 3.4 & 7 & 260/261 \\
Multi-agent & 698,622 & 9,628 & 39.2 & 293.1 & 9.1 & 14 & 258/261 \\
Full System & 650,018 & 9,366 & 35.4 & 290.9 & 8.6 & 14 & 260/261 \\
\midrule
\multicolumn{8}{l}{\textit{FutureX ($N=503$)}} \\
Static Sonnet & 110,392 & 1,435 & 13.4 & 244.4 & 0.0 & 0 & 503/503 \\
A-Evolve & 536,607 & 2,390 & 16.7 & 98.8 & 5.5 & 25 & 501/503 \\
GEPA & 9,233 & 209 & 0.0 & 4.6 & 1.2 & 26 & 503/503 \\
Meta-Harness & 77,662 & 914 & 4.4 & 56.4 & 2.6 & 23 & 503/503 \\
Continual & 57,969 & 761 & 4.6 & 96.9 & 7.3 & 26 & 503/503 \\
SkillOS & 43,830 & 866 & 5.2 & 120.3 & 3.0 & 26 & 503/503 \\
OctoTools & 567,517 & 7,912 & 31.3 & 222.9 & 0.0 & 0 & 498/503 \\
Adaptive & 144,513 & 1,737 & 9.4 & 74.5 & 3.4 & 12 & 500/500 \\
Multi-agent & 263,048 & 2,064 & 11.1 & 86.4 & 21.7 & 21 & 497/503 \\
Full System & 50,867 & 1,006 & 4.0 & 47.6 & 25.0 & 26 & 503/503 \\
\bottomrule
\end{tabular}}
\end{table*}

\noindent\textbf{Statistical procedure.}
Table~\ref{tab:polybench_main} retains its original 1,000-resample marginal
bootstrap intervals (seed 0). Table~\ref{tab:ctf_futurex_main} and the revised
paired analyses use 10,000 resamples with stable per-comparison seeds.
Binary resampling uses equivalent binomial/multinomial counts; paired
resampling preserves joint outcomes on identical tasks. Exact two-sided
McNemar tests use discordant success pairs. Tests are exploratory and
unadjusted unless stated; HITL rates use Wilson intervals.

Table~\ref{tab:headline_pairs} additionally resamples adjacent batch pairs
where task-to-batch assignments agree; a dash denotes incompatible
assignments. All intervals condition on the recorded histories.
This deterministic statistical reconstruction is distinct from rerunning
agents, whose trajectories can vary with hosted APIs and retrieval.

\begin{table*}[htbp]
\centering\footnotesize
\setlength{\tabcolsep}{5pt}
\caption{Paired headline comparisons on identical task IDs. Task-bootstrap CIs condition on the observed runs; the block sensitivity resamples adjacent pairs of recorded batches and also keeps their realized histories fixed. Neither estimates variation from rerunning the evolver. $p$ is a two-sided exact McNemar test, unadjusted across these exploratory comparisons. A non-significant difference does not establish equivalence.}
\label{tab:headline_pairs}
\begin{tabular}{@{}lrrrr@{}}
\toprule
Comparison (first minus second) & $\Delta$ (points), task CI & Batch-block CI & Wins/losses & $p$ \\
\midrule
PolyBench: Full System - Market-following & \makecell{+8.9\\{\scriptsize [7.7, 10.1]}} & [5.6, 13.8] & 728/277 & $<0.001$ \\
PolyBench: Full System - Multi-agent & \makecell{+1.1\\{\scriptsize [0.3, 1.9]}} & [-0.2, 2.4] & 229/175 & 0.008 \\
PolyBench: Full System - Continual & \makecell{+4.0\\{\scriptsize [3.2, 4.7]}} & [1.0, 8.3] & 305/103 & $<0.001$ \\
PolyBench: Adaptive - Continual & \makecell{+0.5\\{\scriptsize [-0.3, 1.3]}} & [-3.3, 4.9] & 248/222 & 0.249 \\
CTF-Dojo: Full System - Multi-agent & \makecell{+2.3\\{\scriptsize [-2.3, 6.5]}} & [-0.8, 6.1] & 20/14 & 0.392 \\
CTF-Dojo: Full System - Continual & \makecell{+3.8\\{\scriptsize [-0.4, 8.0]}} & --- & 21/11 & 0.110 \\
FutureX: Multi-agent - Full System & \makecell{+2.2\\{\scriptsize [-1.2, 5.4]}} & [-0.6, 5.2] & 39/28 & 0.222 \\
FutureX: Multi-agent - A-Evolve & \makecell{+2.0\\{\scriptsize [-1.2, 5.2]}} & [-1.3, 5.4] & 38/28 & 0.268 \\
\bottomrule
\end{tabular}
\end{table*}

\section{Mechanism Analysis: Supporting Evidence}
\label{app:mechanism}

We complement the artifact analyses of the main text with
cross-configuration artifact fingerprints (Table~\ref{tab:mechanism}).
GEPA's evaluated port emphasizes output format and
exhibits zero-search behavior; Meta-Harness retains instructions such as
\emph{``DO NOT SEARCH \ldots\ VERIFIED FACTS ALWAYS OVERRIDE SEARCH RESULTS''}.
These observations characterize the ports: omitted selection or rollback
can affect which policies survive.

\noindent\textbf{Fingerprint definitions.}
\textsc{skip} counts come from final prompts and tool counts from final
workspaces. Keywords alone do not prove contradictions, and file counts
do not measure tool quality or use. The gate flag excludes explicit
\textsc{skip}, so it is not the complement of Coverage. Continual has no
evolved solver prompt, leaving its net-\textsc{skip} entry undefined.

\begin{table}[htbp]
\centering\scriptsize
\setlength{\tabcolsep}{1.8pt}
\caption{\zw{Descriptive artifact fingerprints from the evaluated runs. \textbf{lines}: final solver prompt length; \textbf{net}: \textsc{skip} keyword rules minus never-\textsc{skip} clauses; \textbf{\%gat}: logged gate flags, excluding other abstentions; \textbf{tools}: reusable tool artifacts; \textbf{0-trn}: direct-response proxy from the logged step count. These proxies do not independently measure consolidation, contradiction, search correctness, or tool quality. See Appendix~\ref{app:mechanism}.}}
\label{tab:mechanism}
\begin{tabular}{@{}lrrrr@{\hskip 4pt}rr@{\hskip 4pt}rrr@{}}
\toprule
& \multicolumn{4}{c}{\textbf{PolyBench}} & \multicolumn{2}{c}{\textbf{CTF-Dojo}} & \multicolumn{3}{c}{\textbf{FutureX}} \\
\cmidrule(lr){2-5} \cmidrule(lr){6-7} \cmidrule(lr){8-10}
System & lines & net & \%gat & acc & tools & Pass@1 & tools & 0-trn & Pass@1 \\
\midrule
Static Sonnet & \phantom{0}27 & 1 & 63.5 & 22.2 & 0 & 37.2 & 0 & \phantom{00}0.2 & 31.0 \\
A-Evolve & 367 & 19 & 66.3 & 18.4 & 8 & 45.2 & \textbf{23} & \phantom{00}4.2 & \textbf{47.5} \\
GEPA & 733 & 12 & 69.2 & 13.4 & 0 & 42.9 & 0 & 100.0 & 28.2 \\
Meta-Harness & 540 & 9 & 43.9 & 50.8 & 0 & 41.0 & 1 & \phantom{00}5.6 & 29.4 \\
Continual & seed & --- & \phantom{0}5.4 & 76.9 & \textbf{30} & 46.4 & 0 & \phantom{00}0.0 & 31.0 \\
SkillOS & 264 & 23 & 68.3 & 21.4 & 0 & 29.5 & 0 & \phantom{00}0.8 & 29.8 \\
OctoTools & \phantom{0}73 & 0 & 40.7 & 40.0 & 2 & 36.0 & 0 & \phantom{0}99.2 & 43.3 \\
\midrule
\textbf{Multi-agent} & 161 & $-2$ & \phantom{0}4.1 & 79.8 & \textbf{20} & 47.9 & \textbf{41} & \phantom{00}4.0 & \textbf{49.5} \\
\textbf{Adaptive} & 132 & $-3$ & \phantom{0}8.5 & 77.4 & 8 & 46.0 & 6 & \phantom{00}8.5 & 44.1 \\
\textbf{Full System} & \phantom{0}97 & $-1$ & \phantom{0}1.8 & \textbf{80.9} & 13 & \textbf{50.2} & 1 & \phantom{00}4.0 & 47.3 \\
\bottomrule
\end{tabular}
\end{table}

\noindent\textbf{Illustrative revision.}
Our full system's PolyBench harness adds a scoped tool rule at cycle~28 while retaining
the priority of an existing evidence rule, a question-reading fallback at
cycle~29, and a matching prompt requirement at cycle~33. These edits
coordinate instructions with tool inputs; they illustrate maintenance
without isolating its effect on accuracy.

\section{Metrics and Evidence for D1--D3}
\label{app:stream-dimensions}

Figure~\ref{fig:dimension_correlations} is recomputed offline from saved
per-task results, task metadata, and recorded frozen-harness replays,
without model or retrieval calls; resampling uses fixed seeds, and
repeated builds reproduce every number. Table~\ref{tab:rq3_metrics}
defines the metrics, and Table~\ref{tab:rq3_inputs} gives their
per-stream inputs.

\begin{table*}[t]
\centering
\footnotesize
\setlength{\tabcolsep}{3.5pt}
\caption{D1--D3 metrics in Figure~\ref{fig:dimension_correlations}.
$m$ indexes systems (an evolving configuration or static Sonnet), $t$
tasks, and $\mathcal{B}$ batches.}
\label{tab:rq3_metrics}
\begin{tabular}{@{}>{\raggedright\arraybackslash}p{0.07\textwidth}>{\raggedright\arraybackslash}p{0.22\textwidth}>{\raggedright\arraybackslash}p{0.6\textwidth}c@{}}
\toprule
 & Metric & Definition & Panels \\
\midrule
D1 & Effective coverage $c_{mt}\in\{0,1\}$ & $1$ if system $m$ answers task $t$ through the task's required procedure (Table~\ref{tab:rq3_inputs}), from the saved result & a, d \\
 & Transition & \emph{gained} ($c_{\mathrm{static},t}{=}0$, $c_{mt}{=}1$), \emph{kept} ($1,1$), \emph{lost} ($1,0$), \emph{neither} ($0,0$) & a, d \\
D2 & Heterogeneity $h_t\in[0,1]$ & mean Gower distance from $t$ to all tasks of earlier batches over solver-visible properties (Eqs.~\ref{eq:gower}--\ref{eq:het}) & b, e \\
D3 & Shift $E(\mathcal{B},\mathcal{R})$ & energy distance between the tasks of batch $\mathcal{B}$ and a reference set $\mathcal{R}$ over the same properties; zero when their distributions match (Eq.~\ref{eq:energy}) & c, f \\
 & Reference $\mathcal{R}$ & (c) the $N$ batches before $\mathcal{B}$; (f) the $N$ batches up to a frozen harness's freeze cycle; $N$ from Eq.~\ref{eq:horizon} & c, f \\
Outcome & Gain $g_{mt}$ & $y_{mt}-y_{\mathrm{static},t}$, the accuracy difference on the same task (points) & d, e \\
\bottomrule
\end{tabular}
\end{table*}

\begin{table*}[t]
\centering
\footnotesize
\setlength{\tabcolsep}{3.5pt}
\caption{Per-stream inputs. Task properties are solver-visible and contain
no outcome; level counts are in parentheses, and properties marked
$^{\log}$ are log-scaled. Effective coverage: share of tasks effectively covered by
static Sonnet / by the evolving configurations. $N$: reference window in
batches (Eq.~\ref{eq:horizon}).}
\label{tab:rq3_inputs}
\begin{tabular}{@{}>{\raggedright\arraybackslash}p{0.085\textwidth}>{\raggedright\arraybackslash}p{0.235\textwidth}c>{\raggedright\arraybackslash}p{0.19\textwidth}>{\raggedright\arraybackslash}p{0.262\textwidth}c@{}}
\toprule
Stream & Effectively covered when & \makecell{Effective\\coverage (\%)} & Discrete properties & Continuous properties & $N$ \\
\midrule
PolyBench & a trade is committed, neither gated nor skipped & 32.0 / 65.1 & topic (8), market form (5) & maximum displayed price, bid--ask width, top-3 depth$^{\log}$, liquidity$^{\log}$, volume$^{\log}$, time to resolution$^{\log}$ & 4 \\
CTF-Dojo & a flag is submitted that is not a placeholder (empty, \texttt{flag\{unknown\}}, \emph{none}, \emph{n/a}, \ldots) & 77.3 / 89.5 & category (7), Python or output file (2) & year, points$^{\log}$, file count, description length$^{\log}$ & 1 \\
FutureX & at least one tool is called besides answer submission & 99.8 / 85.6 & domain (4), difficulty level (4), Han-script title (2), answer type (3) & option count, prompt length$^{\log}$, forecast horizon$^{\log}$ & 6 \\
\bottomrule
\end{tabular}
\end{table*}

\noindent\textbf{Population.}
An observation pairs one of the eight evolving configurations with static
Sonnet on the same task. We drop the first batch, which the seed harness
solves, and CTF-Dojo and FutureX tasks whose container was lost for either
system, leaving 44{,}265 observations on 5{,}599 tasks
(Table~\ref{tab:rq3_robustness}). Pooled panels weight streams equally;
stream position is the share of a stream's tasks processed before the
task's batch.

\noindent\textbf{D2: heterogeneity.}
Continuous properties $\tilde{x}$ are clipped to their 1st--99th
percentiles and scaled to $[0,1]$ (missing values take the median), so
each of the $P$ properties contributes a difference in $[0,1]$ to the Gower
distance~\cite{gower1971general}:
\begin{equation}
\begin{aligned}
d(i,j)=\frac{1}{P}\Big(&\sum\nolimits_{c}\lvert\tilde{x}_{ic}-\tilde{x}_{jc}\rvert\\
&+\sum\nolimits_{q}\mathbf{1}[x_{iq}\neq x_{jq}]\Big),
\end{aligned}
\label{eq:gower}
\end{equation}
\begin{equation}
h_t=\frac{1}{\lvert\mathcal{H}_t\rvert}\sum_{j\in\mathcal{H}_t} d(t,j),
\label{eq:het}
\end{equation}
where $c$ and $q$ range over continuous and discrete properties and
$\mathcal{H}_t$ holds the tasks of all earlier batches. $h_t$ grows as an
incoming task differs from everything seen before.

\noindent\textbf{D3: shift.}
A stream is stationary when its task distribution does not change over
time. Shift compares the task distributions of two sets with the energy
distance~\cite{szekely2013energy} on $d$:
\begin{equation}
E(A,B)=2\,\bar{d}(A,B)-\bar{d}(A,A)-\bar{d}(B,B),
\label{eq:energy}
\end{equation}
where $\bar{d}$ averages $d$ over pairs (distinct pairs within a set). The
within-set terms remove each set's own spread: $E$ is near zero when $A$
and $B$ come from one distribution, however heterogeneous, and grows as
the distributions part. The same statistic underlies energy-based
change-point detection~\cite{matteson2014nonparametric}.

\noindent\textbf{Reference window.}
$N$ is each stream's stationarity horizon, set from task properties alone.
With the distributional variogram
$\gamma(k)=\mathrm{mean}_b\,E(\mathcal{B}_b,\mathcal{B}_{b-k})$,
\begin{equation}
N=\max\big\{k:\ 0\in\mathrm{CI}_{95}[\gamma(j)-\gamma(1)]\ \ \forall j\le k\big\},
\label{eq:horizon}
\end{equation}
where the paired difference is averaged over batches $b$ and its interval
comes from a moving-block bootstrap~\cite{kunsch1989jackknife} (block
length $j$). Within $N$ batches the stream is approximately stationary, so
a large $E$ marks departure from the current regime. $\gamma$ rises with
lag in every stream, from $0.064$ to $0.298$, $0.037$ to $0.165$, and
$0.044$ to $0.135$; $N$ is identical across bootstrap seeds.

\noindent\textbf{Frozen-harness replays.}
Each replay copies the continual history up to cycle $k$ and then fixes
the harness (Table~\ref{tab:rq3_replays}). A later batch $b>k+1$ gives the
point $\big(E(\mathcal{B}_b,\mathcal{R}_k),\,\mathrm{acc}_k(b)-
\mathrm{acc}_{\mathrm{cont}}(b)\big)$, with $\mathcal{R}_k$ the $N$
batches up to $k$. This is panel~(c)'s measure with the reference held at
the freeze point, so it also accumulates the drift since $k$: median
$0.165$, versus $0.036$ for the continually evolved harness.

\begin{table}[t]
\centering
\footnotesize
\setlength{\tabcolsep}{2.6pt}
\caption{Frozen-harness replays in Figure~\ref{fig:dimension_correlations}f
(PB: PolyBench, test batches 41--51; CTF: CTF-Dojo, test batches 10--13).
Loss: mean accuracy against continual evolution (points). $r$: correlation
of loss with distance from the freeze point; within: after removing each
harness's mean.}
\label{tab:rq3_replays}
\begin{tabular}{@{}llcrrrr@{}}
\toprule
 & Config. & Freeze $k$ & Pts & Loss & $r$ & Within \\
\midrule
PB & Multi-agent & 6, 7, 22, 40 & 41 & $-7.0$ & $-0.86$ & $-0.71$ \\
PB & A-Evolve & 7, 22 & 22 & $+0.8$ & $-0.64$ & $-0.22$ \\
CTF & Multi-agent & 3, 9 & 7 & $-12.1$ & $+0.33$ & $+0.23$ \\
CTF & A-Evolve & 3 & 4 & $-15.0$ & $-0.95$ & $-0.95$ \\
\midrule
All & & & 74 & $-5.6$ & $-0.53$ & \\
\bottomrule
\end{tabular}
\end{table}

\begin{table}[t]
\centering
\footnotesize
\setlength{\tabcolsep}{3pt}
\caption{Plotted quantity and uncertainty per panel. Intervals are 95\%
percentiles of 2,000 bootstrap replicates resampling the listed clusters
within each stream.}
\label{tab:rq3_panels}
\begin{tabular}{@{}c>{\raggedright\arraybackslash}p{0.55\columnwidth}>{\raggedright\arraybackslash}p{0.3\columnwidth}@{}}
\toprule
Panel & Plotted quantity & Resampled clusters \\
\midrule
(a) & share of gained and lost observations by stream fifth & configurations \\
(b) & batch mean of $h_t$, per stream & none (descriptive) \\
(c) & $E$ against the $N$ preceding batches, per stream & none; permutation test \\
(d) & mean $g_{mt}$ per transition & tasks \\
(e) & mean $g_{mt}$ by within-stream quintile of $h_t$ & tasks \\
(f) & replay points of Table~\ref{tab:rq3_replays}; OLS line & test batches \\
\bottomrule
\end{tabular}
\end{table}

\begin{table}[t]
\centering
\footnotesize
\setlength{\tabcolsep}{3pt}
\caption{Per-stream values behind Figure~\ref{fig:dimension_correlations}
(PB: PolyBench; CTF: CTF-Dojo; FX: FutureX; All: equal-weight pool). Gains in points; gained/lost rows give shares of
tasks (\%) in the first and last stream fifth. D2 trend: Spearman
$\rho$ of batch-mean $h_t$ with stream position. D2 slope: per standard
deviation of the within-stream heterogeneity percentile, with stream
position controlled. Shifted batches: permutation $p<0.05$.}
\label{tab:rq3_robustness}
\begin{tabular}{@{}lrrrr@{}}
\toprule
 & PB & CTF & FX & All \\
\midrule
Observations & 39{,}800 & 837 & 3{,}628 & 44{,}265 \\
Tasks & 4{,}975 & 141 & 483 & 5{,}599 \\
\midrule
Gain where gained & $+83.1$ & $+54.2$ & -- & $+59.1$ \\
Gain where kept & $+6.8$ & $+0.2$ & $+10.2$ & $+5.7$ \\
Gain where lost & $-64.7$ & $-7.5$ & $-2.1$ & $-24.8$ \\
Gained, first fifth & 42.1 & 23.3 & 0.0 & 21.8 \\
Gained, last fifth & 39.9 & 9.3 & 0.0 & 16.4 \\
Lost, first fifth & 7.0 & 2.3 & 16.6 & 8.7 \\
Lost, last fifth & 11.4 & 8.7 & 13.9 & 11.4 \\
\midrule
D2 trend $\rho$ & 0.60 & 0.85 & 0.58 & -- \\
D2 slope & $-7.2$ & $+6.2$ & $-3.0$ & $-1.3$ \\
\midrule
Shifted batches (\%) & 96 & 83 & 38 & -- \\
\bottomrule
\end{tabular}
\end{table}

\noindent\textbf{Robustness.}
Table~\ref{tab:rq3_robustness} reports each stream. The main checks:
\begin{compactitem}
\item \emph{D1 without FutureX}, where static Sonnet already searches on
$99.8\%$ of tasks: gained and lost effective coverage carry $+68.7$
$[57.9,77.4]$ and $-36.1$ $[-42.2,-32.2]$ points; gains fall from
$32.7\%$ to $24.6\%$ of tasks while losses rise from $4.7\%$ to $10.1\%$.
\item \emph{D2 per stream}: slopes of $-7.2$ $[-8.5,-5.9]$, $+6.2$
$[1.0,11.4]$, and $-3.0$ $[-6.0,-0.1]$ are individually clear but of
opposite signs; the pooled task-level correlation is $r=-0.06$
$[-0.12,0.00]$.
\item \emph{D3 beyond difficulty and learning}: pairing an early
($k=6,7$) with a later ($k=22,40$) PolyBench Multi-agent harness on the
same batch, the older harness's extra loss rises with its extra distance
($r=0.81$--$0.88$; $0.82$--$0.92$ with the batch's static accuracy
partialled out; ten batches each).
\end{compactitem}

\noindent\textbf{Scope.}
The panels describe recorded histories and do not isolate causes. FutureX
remains exploratory (\S\ref{sec:setup}), and replay evidence is
concentrated in PolyBench (63 of 74 points). CTF-Dojo's one-batch window
leaves 20-task comparisons whose shift estimates are unreliable
(split-half reliability below zero), and the permutation tests in~(c)
treat tasks within a batch as exchangeable.

\section{Task Streams and Capability Diagnostics}
\label{app:nonstationarity}

\subsection{Accumulation and Participation}

\noindent\textbf{Participation behind a negative accumulation trajectory.}
In Figure~\ref{fig:q1_dynamics}, SkillOS's stored PolyBench instructions grow
$107.8\%$ from stages~2 to~4, while its answering rate drops from $47.2\%$
to $3.5\%$ and its same-task advantage changes from $+19.6$ to $-20.7$
points. Final-stage abstention loses 248 tasks answered correctly by
static Sonnet, offset by five newly correct answers; jointly answered
tasks contribute zero net change. This decomposes the observed deficit
without attributing it to individual updates.

\noindent\textbf{Accumulation measurements.}
Figure~\ref{fig:q1_dynamics} retains all eight evolving configurations
on 5,075 PolyBench and 261 CTF-Dojo tasks. Size sums UTF-8 bytes in the
active main-harness prompt, skills, and serialized memory, excluding
tool code and bookkeeping artifacts. Stage~0 is the seed; later points
use the snapshot preceding each stage's final scored batch.

Stages divide histories at 25/50/75\% task exposure, assigning each
whole batch to its starting stage. PolyBench stage sizes are
1,300/1,300/1,300/1,175; CTF-Dojo sizes are 80/60/60/61, except
SkillOS's resumed schedule (69/80/60/52). Gains use static Sonnet on
identical task IDs. All 16 histories and 64 stages remain, including
negative gains and intervals without updates; A-Evolve stops updating
on PolyBench after cycle~20 (2,000 tasks).

Maintained text need not equal solver input: the loader reads designated
skill documents and memory-content fields, while other layouts can add
bytes without entering that context. Complete historical provider inputs
are not reconstructed, and tool-based access remains possible.

\begin{figure}[t]
\centering
\includegraphics[width=\linewidth]{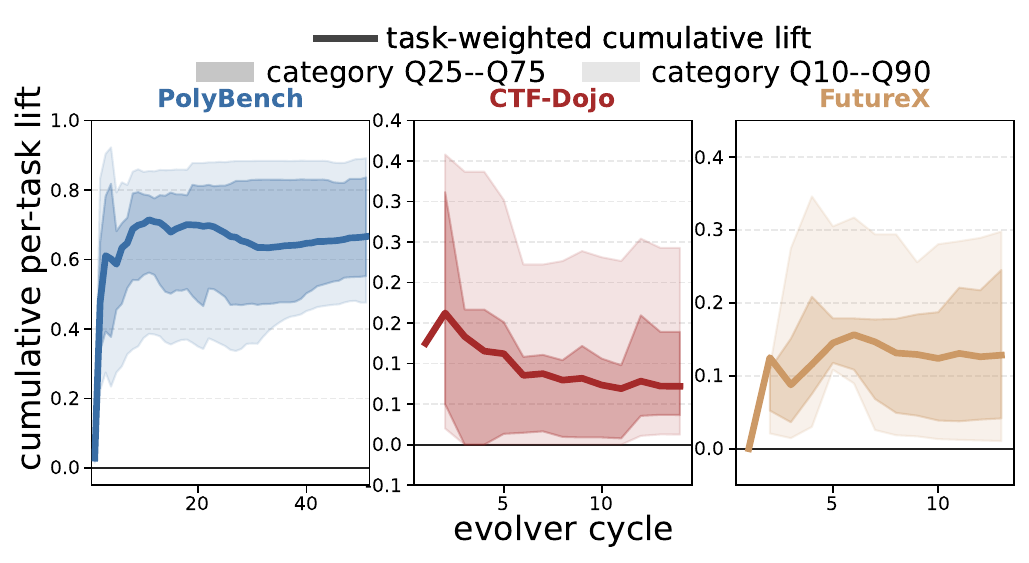}
\caption{\zw{Descriptive adaptive-system lift over the baseline median across cycles. Shaded bands summarize variation across task categories, not confidence intervals or an estimate of $\Ladapt$.}}
\label{fig:l_adapt_evidence}
\end{figure}

Figure~\ref{fig:l_adapt_evidence} complements these histories with
adaptive-system lift over the single-harness baseline median.
Construction also differs between configurations; routing-specific
diagnostics appear in Appendix~\ref{app:branch_perf}.

\subsection{Task-Input Profiles}
\begin{figure}[!t]
\centering
\includegraphics[width=\columnwidth]{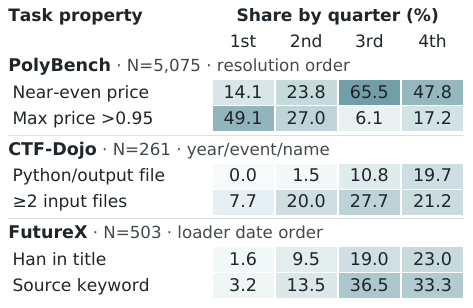}
\caption{\textbf{Task properties change along the retained streams.}
Cells show input-property prevalence in successive task-count quarters;
features may overlap. Quarters can split batches, unlike the whole-batch
stages in Figure~\ref{fig:q1_dynamics}. Definitions appear below.}
\label{fig:task_dimensions}
\end{figure}

Figure~\ref{fig:task_dimensions} aligns prices, listed input files, and
question text to every task in the three main cohorts. Quarters divide
each cohort at $\lfloor qN/4\rfloor$, $q=0,\ldots,4$, and may split
batches. Features can overlap; each denominator includes all tasks
in its quarter. Neither answers nor solver outcomes enter these profiles.

\noindent\textbf{Feature definitions and ordering.}
For PolyBench, near-even means $0.4<\max(\text{displayed prices})<0.6$;
high-price means the maximum exceeds 0.95; ordering follows resolution
time (Appendix~\ref{app:benchmarks}). For CTF-Dojo, Python/output marks a
listed Python file or a filename containing ``output''; the other feature
counts at least two files. Tasks follow year, event, and name order.
For FutureX, Han-title presence matches Unicode U+4E00--U+9FFF.
Source keywords match title or prompt against the archived platform and
finance dictionary, including ``maoyan'', ``flixpatrol'', and ``stock
market capitalization''. These are text cues, not retrieval-access tests.
Ordering retains the loader's estimated dates and malformed-date fallback.

The detailed profiles use complementary partitions:
resolution-date windows for PolyBench, an 80-task cryptography subset
for CTF-Dojo, and the archived exploratory FutureX grouping.
The latter's historical row join differs from the current 503-task cohort;
Figure~\ref{fig:task_dimensions} supplies the aligned main-cohort profile.

\begin{figure*}[t]
\centering
\includegraphics[width=0.9\linewidth]{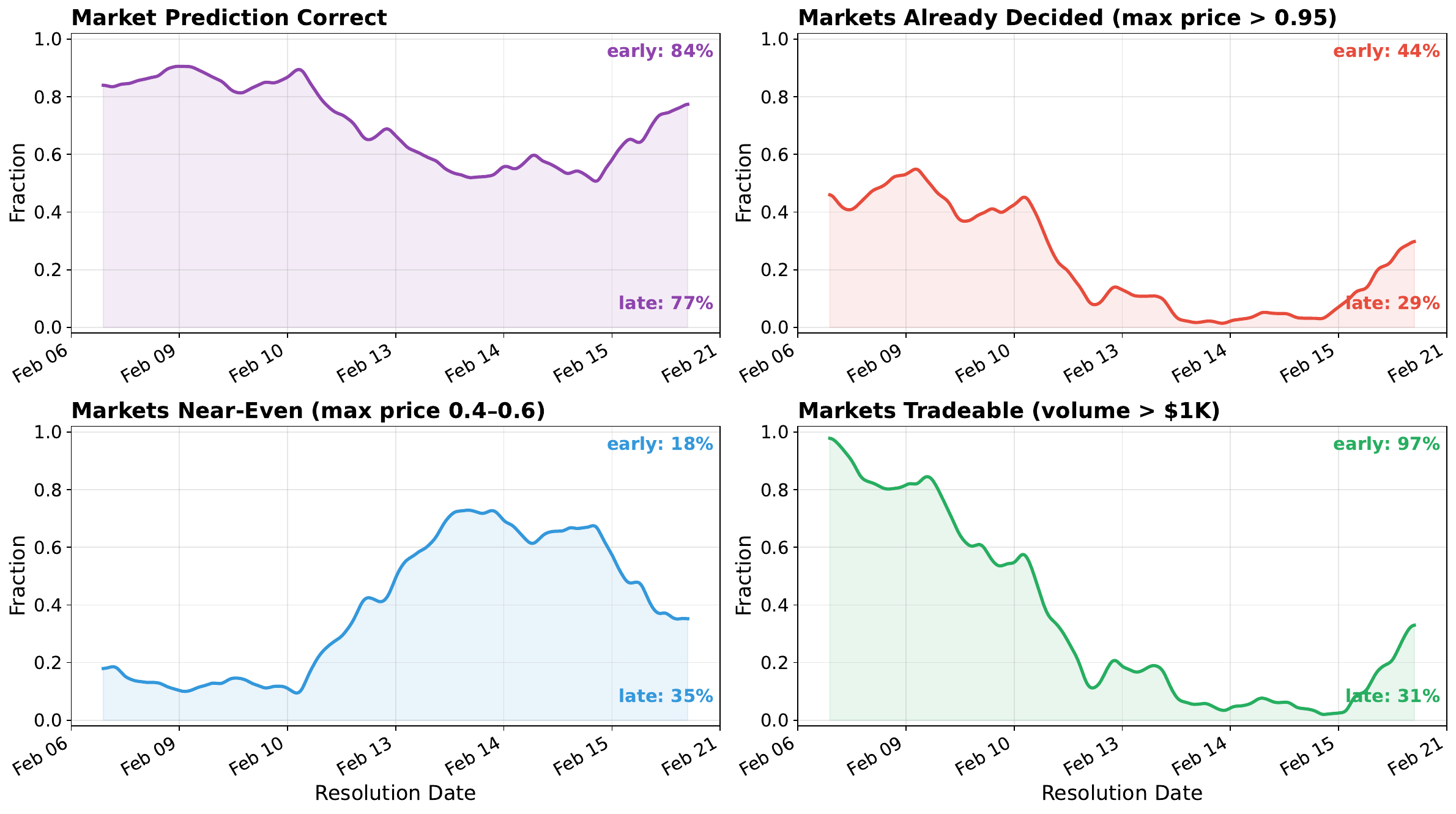}
\caption{\textbf{PolyBench non-stationarity.} Market difficulty and tradability shift over time: later markets are less often decisive or liquid and more often near-even.}
\label{fig:app_nonstationarity_polybench}
\end{figure*}

\begin{figure*}[htb]
\centering
\includegraphics[width=0.8\linewidth]{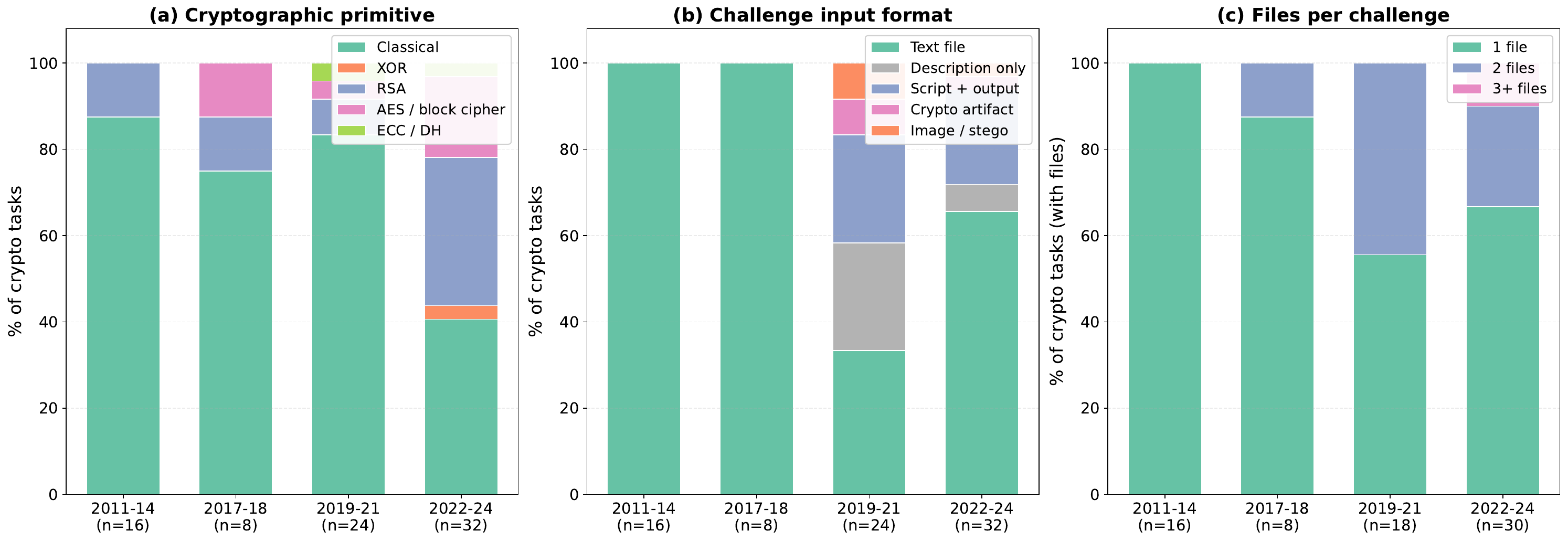}
\caption{\textbf{Archived CTF cryptography profiles.}
Primitive and input-format panels cover 16/8/24/32 catalog tasks in
the four displayed periods; file-count panels use the 16/8/18/30
tasks with files. These task-property profiles describe the 80-task
cryptography subset, rather than the complete 261-task evaluation.}
\label{fig:app_nonstationarity_ctf_dojo}
\end{figure*}

\begin{figure*}[htb]
\centering
\includegraphics[width=0.8\linewidth]{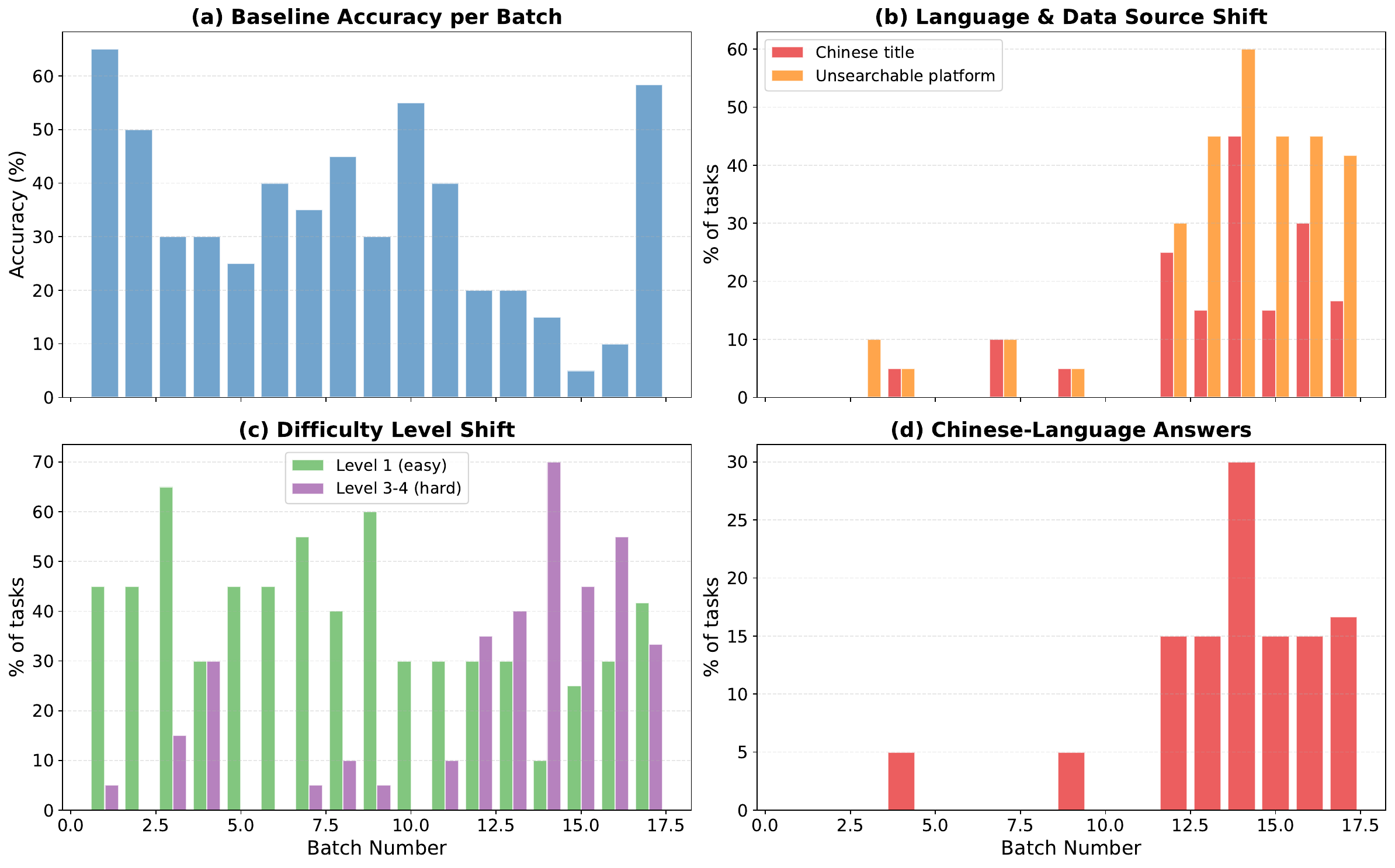}
\caption{\textbf{Archived FutureX exploratory profile.}
The original panels summarize language, source, difficulty, and
answer-format proxies. Their historical row join differs from the
current main cohort; Figure~\ref{fig:task_dimensions} provides the
task-ID-aligned input profile used in RQ1.}
\label{fig:app_nonstationarity_futurex}
\end{figure*}

\subsection{Capability and Workspace Diagnostics}
\label{app:further-experiments}
\label{app:capability-diagnostics}

\begin{figure}[!htbp]
\centering
\includegraphics[width=\columnwidth]{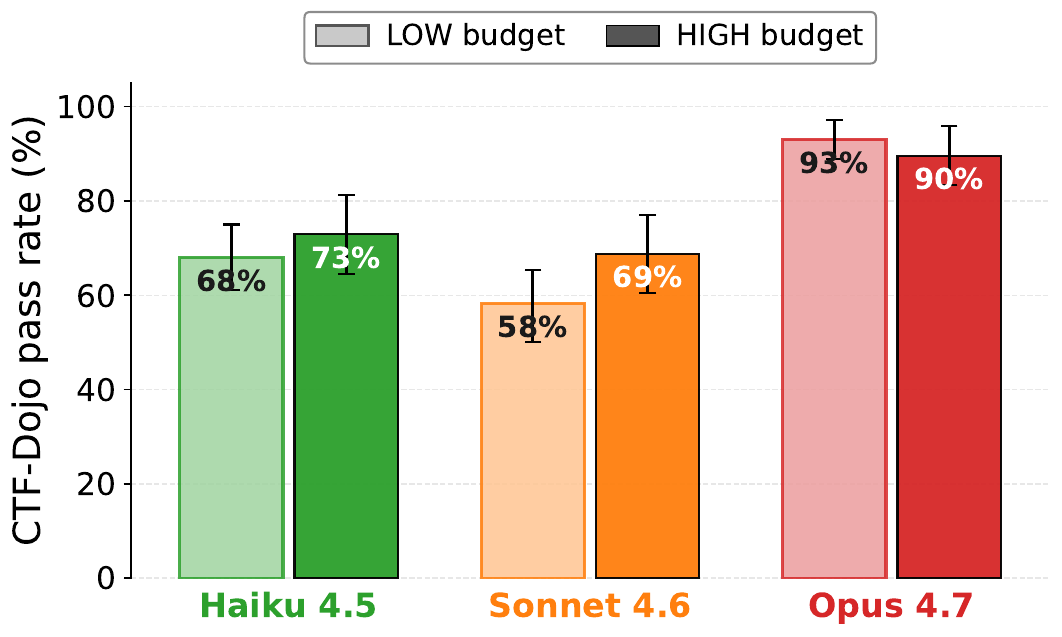}
\caption{\textbf{Evolver capability on CTF-Dojo.} Opus has the highest Pass@1 at both budgets. Higher budget improves Haiku and Sonnet, but not Opus in this separate diagnostic.}
\label{fig:app_c2_evolver_budget}
\end{figure}

\begin{figure}[!htbp]
\centering
\includegraphics[width=\columnwidth]{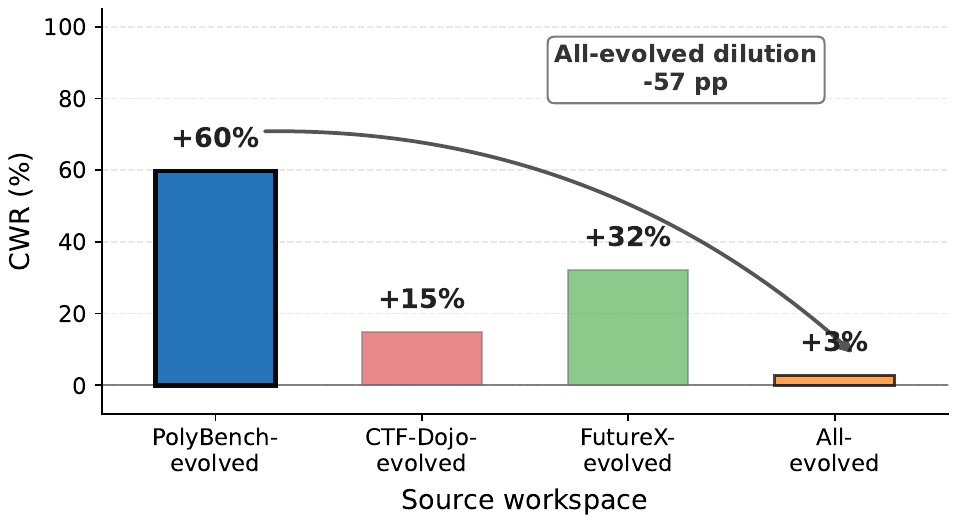}
\caption{\textbf{PolyBench workspace dilution.} The combined workspace scores 57 CWR points below the PolyBench-specific workspace, motivating compatibility checks for accumulated content.}
\label{fig:app_polybench_dilution}
\end{figure}

\noindent\textbf{Capability diagnostics.}
Figure~\ref{fig:l_evo_evidence} uses separate subsets: market consensus
proxies uncertainty rather than outcome calibration; FutureX's
$34.0/47.6/57.1\%$ access-group accuracies also differ in task composition
and temporal exposure. These associations identify construction priorities,
not estimates of $\Levo$. Figures~\ref{fig:app_c2_evolver_budget}
and~\ref{fig:app_polybench_dilution} examine evolver capacity and
workspace compatibility under separate diagnostic conditions.

\noindent\textbf{Failures and budget saturation.}
Tables~\ref{tab:failure_taxonomy} and~\ref{tab:run_capped} give failure categories
and budget-hit counts. Our full system reaches the CTF-Dojo cap on 42.9\%
of tasks; this describes stopping behavior, not the causal benefit of
longer attempts.

\begin{table}[htbp]
\centering\footnotesize
\setlength{\tabcolsep}{5pt}
\caption{Failure taxonomy of our full system. Categories are author-coded from the structured \texttt{timed\_out}, \texttt{max\_turns\_hit}, \texttt{error}, and \texttt{submitted} fields, applied in order so each task is assigned exactly one category. Cells show task count with the percentage of the benchmark in parentheses.}
\label{tab:failure_taxonomy}
\resizebox{\columnwidth}{!}{%
\begin{tabular}{@{}lrrr@{}}
\toprule
Failure mode & PolyBench & CTF-Dojo & FutureX \\
\midrule
Sandbox timeout & 0 & 27\,(10.3\%) & 0 \\
Turn-budget hit & 0 & 85\,(32.6\%) & 0 \\
Sandbox / infra error & 14\,(0.3\%) & 1\,(0.4\%) & 0 \\
No submission & 1\,(0.0\%) & 8\,(3.1\%) & 203\,(40.4\%) \\
Wrong answer & 955\,(18.8\%) & 9\,(3.4\%) & 62\,(12.3\%) \\
\midrule
Total failures & 970\,(19.1\%) & 130\,(49.8\%) & 265\,(52.7\%) \\
Total tasks & 5{,}075 & 261 & 503 \\
\bottomrule
\end{tabular}}
\end{table}

\begin{table}[htb]
\centering\footnotesize
\setlength{\tabcolsep}{4pt}
\caption{Fraction of tasks that hit the per-task turn budget (\texttt{max\_turns\_hit}). High rates on CTF-Dojo signal payload-handling difficulty: a solver that loops on a large binary often exhausts the budget before submitting.}
\label{tab:run_capped}
\resizebox{\columnwidth}{!}{%
\begin{tabular}{@{}lrrr@{}}
\toprule
System & PolyBench & CTF-Dojo & FutureX \\
\midrule
Static Sonnet & 0 & 104\,(39.8\%) & 0 \\
A-Evolve & 0 & 33\,(12.6\%) & 34\,(6.8\%) \\
GEPA & 0 & 92\,(35.2\%) & 0 \\
Meta-Harness & 0 & 99\,(37.9\%) & 2\,(0.4\%) \\
Continual & 0 & 3\,(1.1\%) & 0 \\
SkillOS & 0 & 52\,(19.9\%) & 0 \\
OctoTools & 0 & 106\,(40.6\%) & 0 \\
Multi-agent & 0 & 95\,(36.4\%) & 18\,(3.6\%) \\
Adaptive & 0 & 64\,(24.5\%) & 4\,(0.8\%) \\
Full System & 0 & 112\,(42.9\%) & 0 \\
\bottomrule
\end{tabular}}
\end{table}

\section{Component and Deployment Diagnostics}

\subsection{Component Comparisons}
\label{app:evolution_dynamics}

Figure~\ref{fig:rq2_stats} compares the original 60/100/80-task
CTF-Dojo/PolyBench/FutureX subsets; PolyBench uses CWR.
Labels denote intended ablations: memory reset was unverified, and later
full-stream complete/no-feedback arms both retain outcome information.
Some arms also differ in inference region. These records therefore support
configuration contrasts rather than isolated component effects.

\subsection{Routing Outcomes and Overhead}
\label{app:branch_perf}

Figure~\ref{fig:rq3_stats} uses separately constructed, category-initialized
harness trees. Table~\ref{tab:branch_perf} gives original assignments on
60/100/80 CTF-Dojo/PolyBench/FutureX tasks; replay alignment leaves 40/80/58
because early tasks lack later-created branches. The router uses task
and branch descriptions and selects specialized branches throughout these
subsets, although \texttt{main} is allowed (Appendix~\ref{app:prompt-router}).
Branch rates reflect different assigned task mixes.

\begin{table}[htbp]
\centering\footnotesize
\setlength{\tabcolsep}{4pt}
\caption{Original \emph{Adapt} assignments on the routing subsets. N routed counts assigned tasks; outcomes are Pass@1 for CTF-Dojo/FutureX and HitRate among traded PolyBench markets. Branch rates reflect different assigned task mixes.}
\label{tab:branch_perf}
\resizebox{\columnwidth}{!}{%
\begin{tabular}{@{}llrr@{}}
\toprule
Bench & Branch & N routed & Pass / HitRate (\%) \\
\midrule
\multicolumn{4}{@{}l}{\emph{CTF-Dojo (60 tasks across 3 batches; 7 harnesses including main)}} \\
& Cryptography & 20 & 50.0 \\
& Reverse engineering & 20 & 30.0 \\
& Exploitation & 12 & 0.0 \\
& Miscellaneous & 4 & 75.0 \\
& Web & 2 & 0.0 \\
& Forensics & 2 & 50.0 \\
\midrule
\multicolumn{4}{@{}l}{\emph{PolyBench (100 tasks across 4 batches; 5 harnesses including main)}} \\
& Sports & 71 & 67.6 \\
& Finance & 14 & 21.4 \\
& Culture & 12 & 33.3 \\
& Politics and world events & 3 & 66.7 \\
\midrule
\multicolumn{4}{@{}l}{\emph{FutureX (80 tasks across 3 batches; 5 harnesses including main)}} \\
& Level 1 & 28 & 53.6 \\
& Level 2 & 30 & 30.0 \\
& Level 3 & 8 & 0.0 \\
& Level 4 & 14 & 14.3 \\
\bottomrule
\end{tabular}}
\end{table}

\noindent\textbf{Available alternatives.}
\emph{Oracle}/\emph{Naive}/\emph{Worst} use best/fixed-\texttt{main}/worst
replay outcomes; \emph{Adapt} uses original routed outcomes
(Table~\ref{tab:routing_stats}). Oracle$-$Naive is finite-tree headroom,
not $\Ladapt$; Adapt--Naive includes original-run/replay variation.

\begin{table}[htbp]
\centering\footnotesize
\setlength{\tabcolsep}{4pt}
\caption{Figure~\ref{fig:rq3_stats} means with archived 95\% task-bootstrap CIs. Oracle/Naive/Worst use best/fixed-\texttt{main}/worst replay scores; Adapt uses original routed outcomes. PolyBench uses simulated CWR (Appendix~\ref{app:benchmarks}). Headroom is within the replayed tree, not an estimate of $\Ladapt$.}
\label{tab:routing_stats}
\resizebox{\columnwidth}{!}{%
\begin{tabular}{@{}lccc@{}}
\toprule
 & CTF-Dojo & PolyBench & FutureX \\
 & (Pass@1) & (CWR\%) & (Pass@1) \\
\midrule
N tasks & 40 & 80 & 58 \\
\midrule
Oracle & 55.0\,[40.0, 70.0] & $+$12.0\,[$+$2.1, $+$21.2] & 46.6\,[32.8, 60.3] \\
Adapt  & 35.0\,[20.0, 50.0] & $+$5.9\,[$-$6.6, $+$17.4]  & 34.5\,[22.4, 46.6] \\
Naive  & 17.5\,[7.5, 30.0]  & $+$3.2\,[$-$7.5, $+$13.0]  & 39.7\,[27.6, 51.7] \\
Worst  & 7.5\,[0.0, 17.5]   & $-$10.6\,[$-$25.9, $+$4.0] & 22.4\,[12.1, 32.8] \\
\midrule
Oracle$-$Naive & $+$37.5 & $+$8.8 & $+$6.9 \\
Adapt$-$Naive & $+$17.5 & $+$2.7 & $-$5.2 \\
\bottomrule
\end{tabular}}
\end{table}

\noindent\textbf{Errors and overhead.}
Table~\ref{tab:router_errors} distinguishes failures shared by all available
replays from live failures with a successful alternative (12--20\%).
The latter are potential selection errors because executions can vary.
Later probes use saved tasks and final-tree descriptions with temperature~0,
a 256-output-token cap, and no tools. Median routing latency is
1.66--1.96 seconds, excluding construction and solving.

\begin{table}[htbp]
\centering\footnotesize
\setlength{\tabcolsep}{4pt}
\caption{Routing outcomes and call overhead. Potential error: live failure with an available replay success. Alignment requires \texttt{main} and at least one branch replay, giving FutureX 60 tasks versus 58 in Table~\ref{tab:routing_stats}. Bottom: median usage of later routing-call probes on saved tasks and final-tree descriptions.}
\label{tab:router_errors}
\begin{tabular}{@{}lrrrr@{}}
\toprule
Bench. & $N$ & Live success & No replay win & Pot. error \\
\midrule
CTF-Dojo  & 40 & 35\% & 45\% & 20\% \\
PolyBench & 80 & 64\% & 19\% & 18\% \\
FutureX   & 60 & 33\% & 55\% & 12\% \\
\bottomrule
\end{tabular}

\vspace{0.6em}

\begin{tabular}{@{}lrrrr@{}}
\toprule
Routing call & $N$ & In tok & Out tok & Latency (s) \\
\midrule
CTF-Dojo  & 59 & 455   & 46 & 1.68 \\
PolyBench & 100 & 2{,}356 & 55 & 1.96 \\
FutureX   & 80 & 585   & 39 & 1.66 \\
\bottomrule
\end{tabular}
\end{table}

\subsection{Artifact Selection}
\label{app:adaptation}

Table~\ref{tab:adaptation_spectrum} compares loading the whole catalog, dense
retrieval, and retrieval with an LLM reranker on the same 261 CTF-Dojo and
503 FutureX tasks. Retrieval changes accuracy by $+0.8/-0.2$ points;
CTF reranking loses $6.9$ points, with an unadjusted $p=0.036$ that does
not survive Holm correction over four comparisons. Selected fractions
(7.5--25.8\%) measure catalog content, not token or cost savings.
PolyBench has no distinct comparison because its catalog contains one item.

\begin{table}[!htbp]
\centering\footnotesize
\setlength{\tabcolsep}{3pt}
\caption{Selection over a fixed artifact catalog. Pass@1 and $\Delta$ versus the whole catalog have 95\% task-bootstrap CIs; $p$ is exact two-sided McNemar, unadjusted across four comparisons. Selected is the fraction of catalog items chosen, not token or cost savings.}
\label{tab:adaptation_spectrum}
\begin{tabular}{@{}lrrrr@{}}
\toprule
Operator & Pass@1 (\%) & $\Delta$ (points) & $p$ & Selected \\
\midrule
\multicolumn{5}{@{}l}{\textit{CTF-Dojo}} \\
Whole catalog & \makecell{39.1\\{\scriptsize [33.3, 45.2]}} & --- & --- & 1.00 \\
Retrieval & \makecell{39.8\\{\scriptsize [34.1, 46.0]}} & \makecell{+0.8\\{\scriptsize [-5.0, 6.5]}} & 0.896 & 0.26 \\
Reranker & \makecell{32.2\\{\scriptsize [26.4, 37.9]}} & \makecell{-6.9\\{\scriptsize [-13.0, -0.8]}} & 0.036 & 0.21 \\
\midrule
\multicolumn{5}{@{}l}{\textit{FutureX}} \\
Whole catalog & \makecell{30.0\\{\scriptsize [26.0, 34.0]}} & --- & --- & 1.00 \\
Retrieval & \makecell{29.8\\{\scriptsize [26.0, 33.8]}} & \makecell{-0.2\\{\scriptsize [-2.4, 2.0]}} & 1.000 & 0.19 \\
Reranker & \makecell{28.6\\{\scriptsize [24.7, 32.6]}} & \makecell{-1.4\\{\scriptsize [-3.6, 0.8]}} & 0.265 & 0.08 \\
\bottomrule
\end{tabular}
\end{table}

\section{Human Steering Protocol}
\label{app:hitl_log}

\begin{table}[!htbp]
\centering\footnotesize
\setlength{\tabcolsep}{2pt}
\caption{HITL Pass@1 (\%) with Wilson 95\% intervals. R0--R4 cover all 85 tasks; \emph{Post} uses the same 65 post-message tasks per arm. Required source: R0 none; R1 open web; R2 public API; R3 specialist site; R4 non-English or access-gated. Specified and Generic evolve with credentials; Static is frozen and keyless.}
\label{tab:hitl_tier}
\begin{tabular}{@{}lrccc@{}}
\toprule
Tier & $N$ & Static & Specified & Generic \\
\midrule
R0 & 20 & 70.0\,{\scriptsize [48.1, 85.5]} & 60.0\,{\scriptsize [38.7, 78.1]} & 90.0\,{\scriptsize [69.9, 97.2]} \\
R1 & 20 & 60.0\,{\scriptsize [38.7, 78.1]} & 45.0\,{\scriptsize [25.8, 65.8]} & 50.0\,{\scriptsize [29.9, 70.1]} \\
R2 & 20 & 65.0\,{\scriptsize [43.3, 81.9]} & 75.0\,{\scriptsize [53.1, 88.8]} & 75.0\,{\scriptsize [53.1, 88.8]} \\
R3 & 19 & 42.1\,{\scriptsize [23.1, 63.7]} & 68.4\,{\scriptsize [46.0, 84.6]} & 73.7\,{\scriptsize [51.2, 88.2]} \\
R4 & 6 & 0.0\,{\scriptsize [0.0, 39.0]} & 16.7\,{\scriptsize [3.0, 56.4]} & 16.7\,{\scriptsize [3.0, 56.4]} \\
\midrule
Post & 65 & 55.4\,{\scriptsize [43.3, 66.8]} & 61.5\,{\scriptsize [49.4, 72.4]} & 73.8\,{\scriptsize [62.0, 83.0]} \\
\bottomrule
\end{tabular}
\end{table}

\noindent\textbf{Protocol and timing.}
The 85-task FutureX probe uses author-written messages and a fixed response
handler, with no recruited users. All arms share the seed; Static is frozen
and keyless, while steered arms use Sonnet~4.6/Opus~4.6. Each steered arm
records two credential requests (Exa and Serper) before the steer.
One message follows the first 20 tasks; four later reviews receive
\texttt{skip}.
The message supplies source families (\emph{Specified}) or general retrieval
advice (\emph{Generic}). Batch sizes are 20/20/20/20/5.

\noindent\textbf{Controls and uncertainty.}
Table~\ref{tab:hitl_tier} separates the 85-task tier breakdown from the 65
post-message tasks. Static/Specified/Generic solve 36/40/48 of the latter.
Specified--Generic is $-12.3$ points, with task CI $[-21.5,-3.1]$ and
batch-block CI $[-22.5,4.0]$. The static comparison also changes credentials,
evolution, and inference region, so it does not isolate steering.
Tiers describe source-access requirements; temporal scope follows
Appendix~\ref{app:benchmarks}.

\onecolumn
\clearpage
\begin{nolinenumbers}
\section{System Prompts}
\label{app:prompts}

The following templates specify the four evolution roles and solve-time
router. Their text is preserved verbatim; curly-brace placeholders are
substituted per benchmark or task regime at runtime. Model, budget, and
network settings appear in Table~\ref{tab:hyperparams}.

\label{app:prompt-router}
\begin{promptlisting}[Solve-time router]
You are a fast task router. For each task you must pick exactly one branch
from the list provided.

Decision procedure:
1. Read each branch's README, focusing on its "When to route here" section.
2. Match the task description against each branch's positive signals
   (categories, keywords, metadata) and negative signals (NOT applicable).
3. If exactly one specialized branch matches strongly → route there with
   high confidence (>= 0.7).
4. If multiple branches partially match or none match clearly → route to
   "main" with confidence reflecting your uncertainty.
5. Branches without a clear "When to route here" section should be treated
   as low-confidence candidates — prefer main unless the regime is obvious.

Output JSON only:
{"branch": "main", "confidence": 0.8, "reason": "brief reason"}
\end{promptlisting}

\label{app:prompt-researcher}
\begin{promptlisting}[Researcher]
You are a research agent — PHASE 2 of 4 in the evolution cycle.

PHASE SEQUENCE:
  1. ANALYZE  → analyst identified gaps, wrote task_board.md
  2. RESEARCH (you) → discover solutions for your assigned regime
  3. BUILD    → builder reads YOUR research to write code
  4. VERIFY   → verifier tests what the builder created

UPSTREAM: The analyst assigned you regime "{regime}" because tasks
are failing in that area.

DOWNSTREAM: The builder will read your research records to decide
what code to write. For each approach you test, document:
- How to implement it (endpoint, arguments, response format)
- What it covers and what it doesn't
- Whether it works reliably (tested with real calls)

RESEARCH APPROACH:
1. Read /solver_workspace/ to see what's already implemented
2. Read /evolver_workspace/research_log.jsonl to avoid retesting
3. Search the web for solutions, libraries, APIs, reference code
4. Test NEW approaches that complement what already exists

You can utilize Full network access if that's available.

Write findings to /evolver_workspace/tests/research_{regime}.jsonl.

WORKSPACE LAYOUT:
  /solver_workspace/    — solver workspace (read current code)
  /evolver_workspace/   — evolution state
    research_log.jsonl  — existing records (read to avoid retesting)
    tests/              — write findings HERE
  /trajectories/        — READ-ONLY solver conversations per task

BASH OUTPUT IS CAPPED AT 100 KB PER CALL (first 50 KB + last 50 KB,
middle elided). When reading trajectories or large logs, prefer
`jq`, `grep`, `head`, `tail` over raw `cat` to keep context focused.

{benchmark_context}
\end{promptlisting}

\begin{figure*}[!t]
\centering
\begin{minipage}[t]{0.64\textwidth}
\centering
\includegraphics[width=\linewidth]{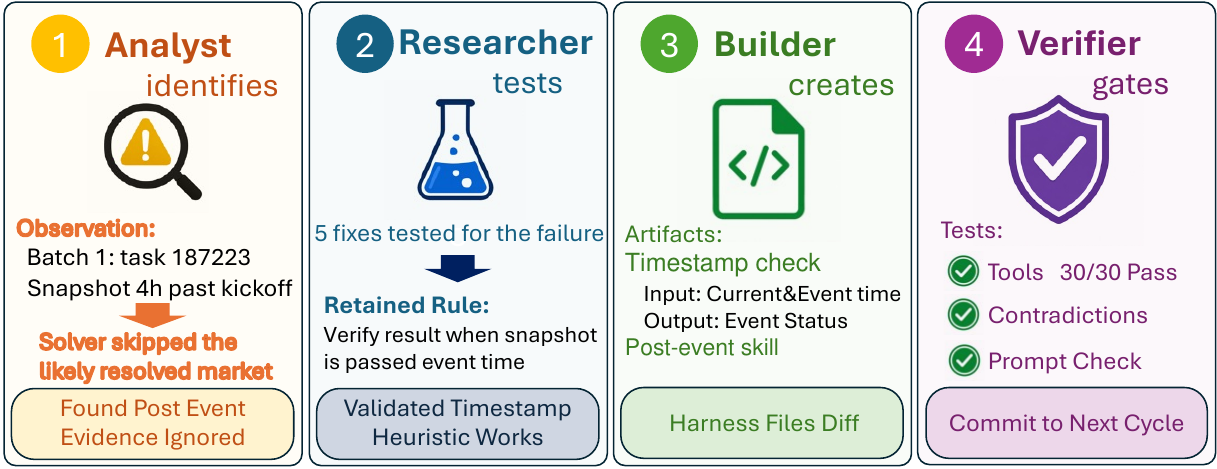}\\[-0.2em]
{\scriptsize (a) Example of the four-phase evolution cycle.}
\end{minipage}\hfill
\begin{minipage}[t]{0.32\textwidth}
\centering
\includegraphics[width=\linewidth]{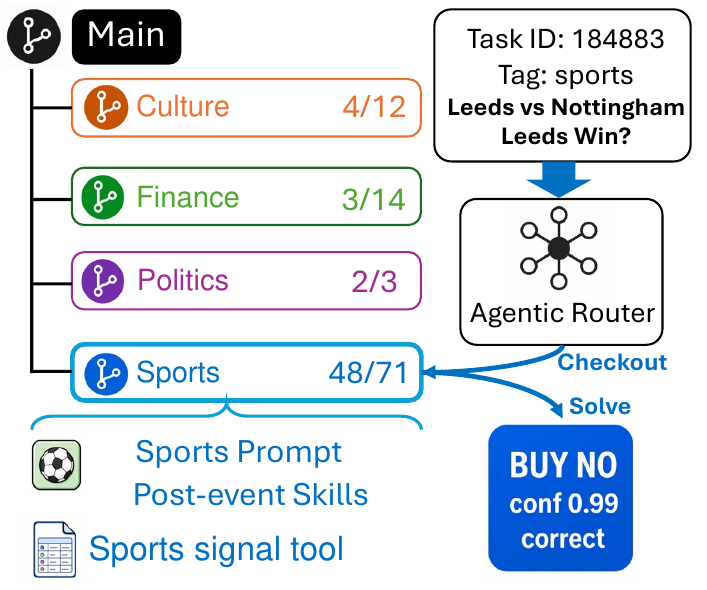}\\[-0.2em]
{\scriptsize (b) Example of the solve-time routing trace.}
\end{minipage}
\vspace{0.25em}

\vskip -0.5em
\caption{Illustrative recorded trajectories of multi-agent evolution and agentic routing. The HITL protocol is documented separately in Appendix~\ref{app:hitl_log}.}
\label{fig:mechanism_traces}
\vspace{-0.4em}
\end{figure*}

\label{app:prompt-builder}
\begin{promptlisting}[Builder]
You are a builder — PHASE 3 of 4 in the evolution cycle.

PHASE SEQUENCE:
  1. ANALYZE  → analyst identified failure patterns in task_board.md
  2. RESEARCH → agents discovered solutions in research_log.jsonl
  3. BUILD (you) → implement solutions from verified research
  4. VERIFY   → verifier tests YOUR code

UPSTREAM: Read research_log.jsonl for verified approaches (works=true).
The task_board tells you which capabilities matter most.
The architecture.md shows what's already built.

DOWNSTREAM: The verifier will test your code. If verification fails,
you get the report and can retry (max 3 attempts).

WORKSPACE LAYOUT:
  /solver_workspace/          — the solver's workspace
    prompts/system.md         — solver prompt (update if needed)
    tools/                    — evolved tool scripts
    infra/                    — evolved infrastructure (if applicable)
{workspace_extras}  /evolver_workspace/
    task_board.md             — failure regimes from analyst
    research_log.jsonl        — verified approaches from research
    architecture.md           — UPDATE with what you built
  /trajectories/              — READ-ONLY solver conversations

Read existing code first. Extend, don't rewrite.
Do NOT run git — the framework handles commits.

BASH OUTPUT IS CAPPED AT 100 KB PER CALL (first 50 KB + last 50 KB,
middle elided). When scanning large files, prefer `jq`, `grep`,
`head`, `tail` over raw `cat`.

{benchmark_context}
\end{promptlisting}

\label{app:prompt-verifier}
\begin{promptlisting}[Verifier]
You are a verification agent — PHASE 4 of 4 in the evolution cycle.

PHASE SEQUENCE:
  1. ANALYZE  → analyst identified failure patterns
  2. RESEARCH → agents discovered solutions
  3. BUILD    → builder wrote code implementing those solutions
  4. VERIFY (you) → test the builder's code before it goes live

UPSTREAM: The builder just wrote/modified code in /solver_workspace/.
Your job is to verify it works before the next batch.

DOWNSTREAM: If PASS, the code goes live. If FAIL, the builder gets
your report and retries. Be specific about what failed.

WORKSPACE LAYOUT:
  /solver_workspace/    — solver workspace (test code here)
  /evolver_workspace/   — evolution state
    task_board.md       — context for test cases
    tests/              — write your test scripts HERE
  /trajectories/        — READ-ONLY solver conversations

Output: VERDICT: PASS or FAIL, then list each test with its result.

BASH OUTPUT IS CAPPED AT 100 KB PER CALL (first 50 KB + last 50 KB,
middle elided). Prefer `jq`, `grep`, `head`, `tail` over raw `cat`
when inspecting large files.

{benchmark_context}
\end{promptlisting}

\label{app:prompt-analyst}
\begin{promptboxcompact}[Analyst]
\begin{multicols}{2}
\begin{lstlisting}[style=promptcompact]
You are a failure analyst — PHASE 1 of 4 in the evolution cycle.

PHASE SEQUENCE:
  1. ANALYZE (you) → write task_board.md with failure patterns + priorities + TARGET
  2. RESEARCH      → agents investigate top-K gaps from YOUR task board
  3. BUILD         → builder implements solutions PER TARGET (main first, then branches)
  4. VERIFY        → verifier tests what the builder created per target

Your task board DIRECTLY drives what gets researched and built next.
Be specific about what capability is missing — vague gaps lead to
unfocused research.

This system uses NAVIGATION: git branches isolate solver strategies.
Each task is routed to the best branch before solving. Your job includes
deciding WHERE each fix should land.

CRITICAL: TRANSFERABILITY ANALYSIS (do this BEFORE writing the task board)

Naive evolution accumulates "shortcut artifacts" — skills, tools, prompt rules,
or memory entries that helped one batch but break later tasks with different
distribution. Examples seen in past runs:
  - A "stop searching at 6 queries" rule helped batch 7 (sports) at 95% but
    crashed batches 14-20 (Chinese music) to 15% because those tasks needed
    longer searches.
  - A `bls_forgery.md` skill with 40s overhead and hard-coded BLS curve
    parameters helped 1 BLS task but slowed all other crypto tasks.
  - 22 hard-coded "wrong flag" entries in system.md memorized past failures
    without abstracting the underlying lesson, bloating the prompt.

Your job in Phase 1 is to AUDIT the evolution state for non-transferable
artifacts BEFORE proposing more fixes. Use bash to:

  1. Compare per-CATEGORY pass rates ACROSS cycles (read multiple
     trajectories/batch_NNNN/index.txt files). If category X had 80% in
     early batches and 30% in recent batches — that's degradation.
  2. List recent additions to skills/, tools/, prompts/system.md, memory/
     and ask: which CATEGORIES does each artifact help vs. hurt?
     - Look for hard-coded task IDs, year-specific logic, single-domain rules
     - Look for prompt rules added recently that contradict older rules
  3. Check the strategy_tree.md routing stats: if a branch is dragging
     overall performance down (worse than main on its routed tasks),
     mark it for retirement.

For each non-transferable artifact found, add a "## Toxic Artifacts" section
to the task board listing:
  - artifact_name: helps {categories} but hurts {categories} → ACTION:
    move-to-branch/<name>, deprecate, or rewrite-as-general

BRANCHING DECISION (TARGET per regime):
- TARGET: main — the fix is domain-generalizable AND verified to NOT degrade
  any previously-passing category
- TARGET: branch/<name> — the fix is regime-specific OR has been observed to
  help one category while hurting another (isolate it from main)
- TARGET: branch/<existing-name> — an existing branch already handles this regime
- Do NOT create new branches for < 2 tasks or single-cycle observations
- Prefer branch isolation when in doubt: keeping main clean is more important
  than maximizing main's per-batch peak

WORKSPACE LAYOUT:
  /solver_workspace/          — the solver's workspace (may be on any branch)
  /evolver_workspace/         — shared evolution state
    task_board.md             — YOUR OUTPUT
    research_log.jsonl        — what's been researched so far
    architecture.md           — what's been built so far
    strategy_tree.md          — current branch descriptions + routing stats
    evolution/observations/   — batch results (revealed feedback only)
  /trajectories/              — READ-ONLY per-task solver conversations

USE BASH to deeply analyze:
- /trajectories/batch_NNNN/index.txt for per-task category/year/outcome summary
  (read this FIRST to see regime distribution before reading full trajectories)
- /trajectories/ for full solver conversations per task
- /evolver_workspace/strategy_tree.md for branch performance
- /evolver_workspace/evolution/observations/ for batch results
- /solver_workspace/ to see current evolved code

BASH OUTPUT IS CAPPED AT 100 KB PER CALL (first 50 KB + last 50 KB,
middle elided). Trajectory JSONs on security/crypto tasks can be
200+ KB each — `cat` will truncate them, and reading several in a row
will exhaust your context. PREFER:
  - `jq '.steps[].tool_use // .steps[].output' traj.json | head -200` to
    see tool calls/outputs without the raw conversation bulk
  - `jq -r '.steps[-5:]' traj.json` to inspect only the final few steps
  - `grep -n ERROR|FAIL|flag traj.json` to locate specific signals
  - `ls -lS /trajectories/ | head` to find the largest trajectories first
  - `wc -l traj.json` before `cat` to check size
Reserve raw `cat` for files under ~50 KB.

NAVIGATION CONTEXT (provided in the user prompt):
- Strategy tree: existing branches and their per-task routing performance
- Routing summary: which tasks went to which branch this batch, and passed/failed

PRIVACY: feedback_archive.jsonl is masked. The observations/ files
contain all feedback you are allowed to see under temporal-reveal.

OUTPUT FORMAT: task board in the exact markdown format with TARGET annotations:
```
## Failure Patterns (Cycle N)
- regime_name: N tasks ... PRIORITY: HIGH → TARGET: main
- other_regime: M tasks ... PRIORITY: MEDIUM → TARGET: branch/regime-name
```

No conversational text in the final output.

{benchmark_context}
\end{lstlisting}
\end{multicols}
\end{promptboxcompact}
\end{nolinenumbers}

\end{document}